\documentclass[letterpaper, 10 pt, journal]{IEEEtran}
\usepackage[colorlinks,urlcolor=blue,linkcolor=blue,citecolor=blue]{hyperref}

\usepackage{color,array}

\usepackage{graphicx}

\usepackage{amsmath,amssymb,amsfonts}
\usepackage{algorithm}
\usepackage{textcomp}
\usepackage{graphicx,color}
\usepackage{mathrsfs}
\usepackage[noend]{algpseudocode}
\usepackage{subfigure}
\usepackage{url}
\usepackage{color}
\usepackage{dsfont}
\usepackage{bbm}
\usepackage{booktabs}
\usepackage{array}
\usepackage[table]{xcolor} 
\usepackage{yfonts}
\usepackage[normalem]{ulem}
\usepackage{amsthm}
\usepackage{arydshln}
\usepackage{accents}
\usepackage{soul}

\allowdisplaybreaks

\newtheorem{theorem}{Theorem}[section]
\newtheorem{lemma}[theorem]{Lemma}

\newtheorem{assumption}[theorem]{Assumption}
\newtheorem{example}{Example}

\usepackage{tabularx}
\usepackage{multirow}
\usepackage{makecell}

\newcommand\aamsout{\bgroup\markoverwith{\textcolor{violet}{\rule[0.5ex]{2pt}{1pt}}}\ULon}

\newcommand{\real}{\mathbb{R}}

\newcommand{\transpose}{\mathsf{T}} 
\newcommand{\T}{\mathsf{T}} 

\newcommand{\mc}{\mathcal}

\newcommand{\lip}{\mathrm{lip}}
\newcommand{\Lip}{\mathrm{Lip}}

\DeclareSymbolFont{bbold}{U}{bbold}{m}{n}
\DeclareSymbolFontAlphabet{\mathbbold}{bbold}

\newcommand{\R}{\mathbb{R}}

\newcommand\oprocendsymbol{\hbox{$\square$}}
\newcommand\oprocend{\relax\ifmmode\else\unskip\hfill\fi\oprocendsymbol}

\DeclareMathOperator{\id}{Id}

\renewcommand{\theenumi}{(\roman{enumi})}

\newcommand*{\QEDA}{\hfill\ensuremath{\blacksquare}}%

\graphicspath{{figs/}}

\begin{document}


\title{\bf Learning Lipschitz Feedback Policies from Expert Demonstrations: Closed-Loop Guarantees, Generalization and Robustness} \author{Abed~AlRahman~Al~Makdah,
  Vishaal~Krishnan, and Fabio~Pasqualetti \thanks{This material is
    based upon work supported in part by awards ONR-N00014-19-1-2264 and
    AFOSR FA9550-20-1-0140. AAAM is with the
    Department of Electrical and Computer Engineering, VK and FP are
    with the Department of Mechanical Engineering at the University of
    California, Riverside,
    \href{mailto:aalm005@ucr.edu}{\{\texttt{aalmakdah}},
    \href{mailto:vishaalk@ucr.edu}{\texttt{vishaalk}},
    \href{mailto:fabiopas@engr.ucr.edu}{\texttt{fabiopas\}@engr.ucr.edu}}.
  } }


%
%



\maketitle
\begin{abstract}
 In this work, we propose a framework to learn
 feedback control policies with guarantees on closed-loop generalization and adversarial robustness.   
  These policies are learned directly from expert demonstrations, 
  contained in a dataset of state-control input pairs,
  without any prior knowledge of the task and system model. 
  We use a Lipschitz-constrained loss minimization scheme
  to learn feedback policies with certified closed-loop robustness,
  wherein the Lipschitz constraint serves as a mechanism to
  tune the generalization performance and robustness
  to adversarial disturbances.  
  Our analysis exploits the Lipschitz property to obtain
  closed-loop guarantees on generalization and robustness
  of the learned policies.
  In particular, we derive a finite sample bound on the policy learning error
  and establish robust closed-loop stability under the learned control policy. 
  We also derive bounds on the closed-loop regret with respect to the expert policy 
  and the deterioration of closed-loop performance under bounded 
  (adversarial) disturbances to the state measurements. 
  Numerical results validate our analysis
  and demonstrate the effectiveness of our robust feedback policy
  learning framework.
  Finally, our results suggest the existence of a
  potential tradeoff between nominal closed-loop performance and
  adversarial robustness, and that improvements in nominal closed-loop
  performance can only be made at the expense of robustness to
  adversarial perturbations.
\end{abstract}


\section{Introduction}\label{sec: introduction}
Robustness of data-driven models to adversarial perturbations has attracted much
attention in recent years. One of the approaches to robust learning
seeks to modulate the Lipschitz constant of the data-driven model~\cite{MF-AR-HH-MM-GJP:19,SA-HG-JC-MU:20,PLC-JCP:20},
either via a regularization~\cite{LB-RR-TR-LS-DT:21,YY-TM:17} of the learning loss function or by
imposing a Lipschitz constraint~\cite{HG-EF-BP-MJC:21,VK-AAALM-FP:20}. Since the Lipschitz constant
determines the (worst-case) sensitivity of a model to perturbations of
the input, data-driven models trained with Lipschitz 
constraints/regularizers are expected to be robust to bounded
(adversarial) perturbations \cite{CS-WZ-IS-JB-DE-IG-RF:14}. Prior works have primarily explored this
approach for static input-output models \cite{VK-AAALM-FP:20,CS-WZ-IS-JB-DE-IG-RF:14,TWW-HZ-PYC-JY-DS-YG-CJH-LD:18}. 
However, in a feedback control setting, a static input-output 
robustness guarantee for a data-driven controller may not result in robust
closed-loop performance. When a data-driven controller is integrated into the feedback loop, 
a static input-output robustness guarantee for the data-driven controller must be combined with
appropriate robust control notions to yield a robustness certificate
for the closed-loop system \cite{FB-MT-AS-AK:17,MJ-JL:20}.
 Obtaining safety and robustness certificates for data-driven controllers in closed-loop systems remains an active area of research in general. In this work, we propose a learning framework to learn Lipschitz feedback policies with provable guarantees on closed-loop performance and robustness against bounded adversarial perturbation, where these policies are learned directly from expert demonstrations without any prior knowledge of the task and the system model.
 
%

The problem of learning optimal feedback control policies from data 
for a nonlinear system with unknown dynamics and control cost is not
only technically challenging, but also has high sample complexity,
which presents obstacles for data collection and the use
of data-driven algorithms. In imitation learning framework, this issue is often mitigated by expert
demonstrations of the optimal feedback policy, which help reduce the
problem to one of learning the policy implemented by the expert. Yet,
learning is not a simple repetition of the expert controls, but rather
the ability to generalize and respond to unseen conditions as the expert demonstrator would. A naive learning approach (such as behavioral cloning in imitation learning \cite{DAP:89}) that overlooks
the generalization and robustness requirements may not only result in pointwise differences
between the expert and implemented policies, but also in
unstable trajectories and failure of the controlled~system \cite{SR-DB:10}. This raises the following question: For an unknown system and control task, 
what is an appropriate method to learn a feedback policy 
from a finite number of expert demonstrations (dataset of state-control input pairs)
such that
(i) the learned policy generalizes expert performance beyond 
the finite data points to a broader region of interest, and
(ii) closed-loop performance remains robust to 
(adversarial) disturbances of the state measurements.

\textbf{Related work.} There have been several proposals to address the above question 
in various settings. Broadly, this problem falls under the 
umbrella of imitation learning, which has been studied extensively in the literature 
and implemented in various contexts including video games~\cite{SR-DB:10,SR-GG-DB:11}, 
robotics~\cite{SS:99} and autonomous driving~\cite{FC-MM-AL-VK-AD:18}.\\ 
\text{\emph{Generalization:}} The key obstacle to widespread adoption of imitation learning is that it is difficult to guarantee performance in unseen scenarios. One approach to overcome this obstacle is inverse reinforcement learning (also referred to as apprenticeship learning in the literature), where the learner infer the unknown cost function from expert demonstrations, then learn an optimal policy that optimizes the learned cost using reinforcement learning \cite{PA-AYN:04,PA-DD-AYN-ST:08,US-RES:08,SL-ZP-VK:11,JH-JG-SE:16}. Since the learned cost represents the task of the expert, inverse reinforcement learning algorithms are able to generalize to unseen scenarios that are not covered by the expert demonstrations. However, one drawback of inverse reinforcement learning is that there can exist multiple cost functions that can be optimized under the expert's policy, which adds ambiguity in learning the cost function \cite{BDZ-AM-JAB-AKD:08}. Another approach that overcomes the obstacle of performing in unseen scenarios is direct policy learning via interactive expert \cite{SR-DB:10,SR-GJG-JAB:11}. In this approach, the learner can query an interactive expert at each iteration, then, the learner uses the expert's feedback to correct its mistakes and improve its policy. Since this approach keeps expanding the expert's data, it will eventually cover all possible scenarios in the long run. However, one drawback of this approach is that it requires the expert to be always available for feedback. In \cite{ML-JL-RF-AD-KG:17}, noise is injected into the expert's policy in order to provide demonstrations on how to recover from errors. In~\cite{PT-AF-RM-NR-SL-YG:19}, the authors develop a framework for learning a generative model for planning trajectories from demonstrations, which allows it to capture uncertainty about previously unseen scenarios. \\
\emph{Closed-loop performance and robustness:} Several approaches to adversarial imitation learning have been proposed in~\cite{IK-KKA-DD-SL-JT:18, KZ-SR-AN-SGC-DB-SC-MD-NDF-ZW:19}, where inverse reinforcement learning is used. In \cite{JW-ZZ-YW-HZ:21}, the authors proposed an adversarially robust imitation learning framework, where an agent is trained in an adversarially perturbed environment with the expert being available for queries at any time step. In \cite{LL-AR-LJ-ST-NM:21}, the authors learn robust control barrier functions from safe expert demonstrations. In all these works, robustness of imitation learning algorithms is considered to be the ability of the learned policy to recover from errors, which is similar to the notion of generalization.

In contrast to many of the works referenced above,
we seek a principled feedback policy learning framework with strong
theoretical guarantees. In particular, we seek explicit bounds on
the finite sample performance, stability and robustness of the
closed-loop system under the learned feedback policy.
The broader problem of obtaining closed-loop performance 
and robustness guarantees for learned 
feedback policies and understanding the underlying tradeoffs 
has attracted attention recently~\cite{AAALM-VK-FP:19b,LH-KW-MM-MZ:20,ST-AR-TZ-NM:21};
yet it remains an active area of research. 
This requires the integration of theoretical tools from
several areas: (i) the underlying control task is typically
specified as an optimal control problem with performance
measured in terms of the cost incurred, 
(ii) the feedback control policy is learned from finite offline data
which involves considerations of generalization and robustness to distributional shifts, and
(iii) closed-loop performance guarantees typically
rely on an underlying robust stability guarantee for the
learned policy.
Prior works have addressed this problem within various frameworks, such as
the $H_{\infty}$-control framework for linear systems~\cite{SD-NM-BR-VY:19}.
However, the problem of obtaining guarantees on closed-loop
generalization and robustness to distributional shifts
of learned policies for general nonlinear systems
still remains a challenge. In this work, we address this problem
within a Lipschitz feedback policy learning framework.
The Lipschitz property is a fairly mild requirement in 
nonlinear control, and through our analysis we will see that 
it can be exploited to provide
closed-loop bounds on generalization and robustness to 
distributional shifts for learned policies, highlighting the
effectiveness of this~approach.

\textbf{Contributions.}  Our primary contribution in this paper is a robust feedback control policy learning framework based on
Lipschitz-constrained loss minimization, where the feedback policies
are learned directly from expert demonstrations.  We then undertake a
systematic study of the performance of feedback policies learned
within our framework using meaningful metrics to measure closed-loop
stability, performance and robustness. Our work integrates
robust learning, optimal control and robust stability into a
unified framework for robust feedback policy learning. 
More specifically, our technical contributions include:
(i) an analysis of the Lipschitz-constrained policy learning
problem, resulting in a finite sample bound on the learning error,
(ii) a robust stability bound for the closed-loop system 
under the learned feedback policies,
(iii) a Lipschitz analysis resulting in a bound on the regret incurred by 
learned feedback policies in terms of the learning error, 
and a bound on the deterioration of performance in the presence of 
(adversarial) disturbances to state measurements. 
This sheds light on the dependence of closed-loop control 
performance and robustness on learning. 
Conversely, our results specify target bounds on policy
learning error for desired closed-loop performance.  
We then demonstrate our robust feedback policy learning framework 
via numerical experiments on
(i) the standard LQR benchmark, and (ii) a non-holonomic differential
drive mobile robot model.
Finally, our analysis points to the existence of a potential tradeoff between 
nominal performance of the learned
policies and their robustness to adversarial disturbances of the
feedback, which is borne out in numerical experiments where we observe
that improvements to adversarial robustness can only be made at the
expense of nominal performance.  

\textbf{Notation.}
For open and bounded sets $\mathcal{X} \subset \real^{{\rm dim}(\mathcal{X})}$
and $\mathcal{Y} \subset \real^{{\rm dim}(\mathcal{Y})}$, 
and a Lipschitz continuous map~$f: \mathcal{X} \rightarrow \mathcal{Y}$, we denote
by~$\ell_{f}$ the Lipschitz constant of~$f$.
Furthermore, we denote by $\Lip(\mathcal{X}; \mathcal{Y})$ 
the space of Lipschitz continuous maps
from $\mathcal{X}$ to $\mathcal{Y}$. 
We denote by $\mathrm{Vol}(\mathcal{X})$ the volume of~$\mathcal{X}$.
A function $g: \mathcal{X} \rightarrow \real$
is said to be $\lambda$-smooth if it has a Lipschitz-continuous gradient, i.e.,
$\left\| \nabla g (x_1) - \nabla g(x_2) \right\| \leq \lambda \left\| x_1 - x_2 \right\|$
for any $x_1, x_2 \in \mathcal{X}$, where~$\| \cdot \|$ denotes the norm operator.
A continously differentiable function $g: \mathcal{X} \rightarrow \real$ is said to be $\mu$-strongly
convex if $\left\| \nabla g (x_1) - \nabla g(x_2) \right\| \geq \mu \left\| x_1 - x_2 \right\|$
for any $x_1, x_2 \in \mathcal{X}$. The maximum eigenvalue of a square matrix $A$ is denoted by $\rho(A)$. The cardinality of a set $\mc{X}$ is denoted by $|\mc{X}|$.

\section{Problem formulation and outline of the approach}	\label{sec:framework_outline}
In the section, we setup the problem of
learning robust feedback control policies from 
expert demonstrations and present an outline of
our approach. 
\subsection{Problem setup}
We begin by specifying the properties of the system, the control task
and the dataset of expert demonstrations.
Consider a discrete-time nonlinear system of the form:
\begin{align}\label{eq: dynamical model}
\begin{aligned}
  x_{t+1} = f(x_t,u_t), \qquad y_t = x_t+\delta_t,
\end{aligned}
\end{align}
where the map $f: \mathbb{R}^n \times \mathbb{R}^m \rightarrow \mathbb{R}^n$ 
denotes the dynamics, $x_t \in \mathbb{R}^n$ the state,
$u_t \in \R^{m}$ the control input
and $y_t \in \mathbb{R}^n$ the full-state measurement 
at time~$t \in \mathbb{N}$, respectively,
with disturbance $\|\delta_t\|\leq \zeta$ 
for any $t \in \mathbb{N}$\footnote{the output equation allows for the modeling of
sensors that are susceptible to bounded 
(adversarial) disturbances~\cite{FP-FD-FB:10y}.}.
\begin{assumption}[\bf \emph{System properties}]
  \label{ass:system}
The following properties hold for System~\eqref{eq: dynamical model}:
\begin{enumerate}
\item {\bf \emph{Fixed point at origin:}}
	The map~$f$ in~\eqref{eq: dynamical model} has a fixed point at the origin (i.e., $f(0,0) = 0$).
\item {\bf \emph{Lipschitz continuous dynamics:}}
	The map~$f$ in~\eqref{eq: dynamical model}
	is Lipschitz continuous with constants~$\ell_f^x$ and~$\ell_f^u$ 
	(i.e., $\left\| f(x_1, u_1) - f(x_2, u_2) \right\| \leq \ell_f^x \| x_1 - x_2 \| + \ell_f^u \| u_1 - u_2 \|$
	for any $x_1, x_2 \in \real^n$ and $u_1, u_2 \in \real^m$).
\item {\bf \emph{Exponential stabilizability by Lipschitz feedback:}}
	System~\eqref{eq: dynamical model} is
	uniformly exponentially stabilizable by Lipschitz feedback, i.e.,
	there exists a Lipschitz continuous feedback policy~$\pi$ 
	and constants $M \in \real_{\geq 0}$, $\beta \in (0,1)$
	such that $\left\| f_\pi^t (x) \right\| \leq M \beta^t \| x \|$.   \oprocend
\end{enumerate}
\end{assumption}
\medskip
\noindent
We now explain the motivation behind the above assumptions on
the properties of System~\eqref{eq: dynamical model}.
The control task is often formulated as one of stabilizing the system
to the origin. Assumption~\ref{ass:system}-(i) states that the origin,
in the absence of control input, is indeed a fixed point of the system.
The Lipschitz continuity Assumption~\ref{ass:system}-(ii) 
specifies the level of regularity intrinsic to the system dynamics
and is fairly standard in the literature.
From a control design perspective it is crucial that the system indeed
possesses the desired stabilizability properties 
from within this class of feedback policies considered in design.
In this paper, we seek to learn feedback policies with Lipschitz regularity and
Assumption~\ref{ass:system} specifies that this is the case 
and that System~\eqref{eq: dynamical model} is exponentially stabilizable by Lipschitz feedback.

The task is one of infinite-horizon 
discounted optimal control of System~\eqref{eq: dynamical model}
by a Lipschitz-continuous feedback policy, 
with stage cost $c : \real^n \times \real^m \rightarrow \real_{\geq 0}$
and discount factor $\gamma \in (0,1)$:
\begin{align}\label{eq:opt_control_value_function}
  \begin{aligned}
   	&\min_{\pi \in \Lip(\real^n; \real^m)}~\sum_{t=0}^{\infty} \gamma^t c(x_t, u_t), \\
   	&\qquad\;~\text{s.t.}~ \begin{cases} 
	x_{t+1} &= f(x_t,u_t), 
	 \\ u_t &= \pi(x_t+\delta_t),
	  \end{cases}
      \end{aligned}
\end{align}
where $\Lip(\real^n; \real^m)$ is the space of Lipschitz-continuous feedback policies.
Furthermore, we would like the closed-loop performance
to be robust to the disturbance~$\delta$.
\begin{assumption}[\bf \emph{Task properties}]\label{ass:task}
The following hold for Task~\eqref{eq:opt_control_value_function} and System~\eqref{eq: dynamical model}:
\begin{enumerate}
\item {\bf \emph{Strong convexity and smoothness of stage cost:}}
 	The stage cost $c: \real^n \times \real^m \rightarrow \real_{\geq 0}$ is
	$\mu$-strongly convex and $\lambda$-smooth. 
	Furthermore, $c(x,u) = 0$ if and only if $x=0$ and $u=0$.
\item {\bf \emph{Existence of optimal feedback policy:}}
  For every $\gamma \in (0,1)$, there exists a minimizer $\pi^* \in \Lip(\real^n; \real^m)$
  to the optimal control problem~\eqref{eq:opt_control_value_function} with $\delta \equiv 0$.    \oprocend
\end{enumerate}
 \end{assumption}
\medskip
\noindent
The choice of optimal control cost function plays an important role in
determining the properties of the optimal feedback policy.
Assumption~\ref{ass:task}-(i) specifies the 
convexity and smoothness properties of the control cost.
Existence of an optimal feedback policy within the considered class, 
as specified in Assumption~\ref{ass:task}-(ii) is a minimum requirement
for control design.

We now verify the properties in Assumptions~\ref{ass:system} and~\ref{ass:task}
in the Linear-Quadratic control setting.
\begin{example}[\bf \emph{Linear quadratic control}]
  For a linear system with~$f(x,u) = Ax + Bu$ such that
  $(A,B)$ is a controllable pair, it can be seen that the properties in
  Assumption~\ref{ass:system} readily follow.
  It can be seen that a quadratic stage cost
  $c(x,u) = x^\top Q x + 2 x^\top W u + u^\top R u$ (with $Q > 0$ and
  $R - W^\top Q^{-1} W > 0$) is strongly convex and has a
  Lipschitz-continuous gradient with $\mu = \lambda_{\min} (H)$,
  $\lambda = \lambda_{\max} (H)$, and
  $H = 2 \left[ \begin{matrix} Q & W \\ W^\top & R \end{matrix}
  \right]$, thereby satisfying Assumption~\ref{ass:task}-(i).
  Furthermore, we note that Assumption~\ref{ass:task}-(ii) readily follows
  from the existence of an optimal feedback gain for the discounted
  infinite-horizon LQR problem, and the fact that the corresponding optimal value function is
  quadratic.
\end{example}

\medskip
\noindent
In this paper, we consider the problem of data-driven
feedback control, where we have access neither to the underlying dynamics~$f$ nor to the task
cost function (stage cost~$c$ and discount factor~$\gamma$).
Instead, we have access to $N < \infty$ expert demonstrations of
an (unknown) optimal feedback policy~$\pi^*$ on System~\eqref{eq:
  dynamical model} over a finite horizon of length~$T$. The initial
state of the demonstrations is sampled uniformly i.i.d. from
$B_r(0) \subset \real^n$, the ball of radius $r$ centered at the
origin. The data is collected in the form of matrices $X, U$ as follows:
\begin{align*} X =\left[ \begin{matrix} \mathbf{x}^{(1)} & \ldots
     & \mathbf{x}^{(N)} \end{matrix} \right], \quad U =\left[
   \begin{matrix} \mathbf{u}^{(1)} & \ldots & \mathbf{u}^{(N)}
   \end{matrix} \right],
   \end{align*}
   where $\mathbf{x}^{(i)} = ( x_0^{(i)}, \ldots, x_T^{(i)} )$ and $\mathbf{u}^{(i)} = ( u_0^{(i)}, \ldots, u_{T-1}^{(i)} )$ are the state and input vectors from the $i$-th demonstration, satisfying $u_t^{(i)} = \pi^* (x_t^{(i)} )$ for all $i \in \{ 1, \ldots, N \}$ and $t \in \{ 0, \ldots, T-1 \}$. 

\subsection{Outline of the approach}
\begin{figure}[t]
  \centering
  \includegraphics[width=1\columnwidth]{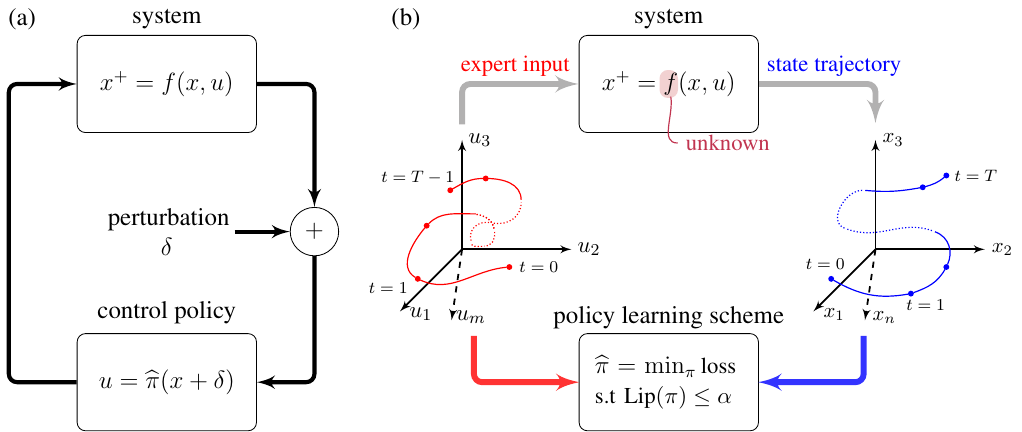}
  \caption{The block diagram in panel (a) corresponds to the 
  implementation of the learned control policy~$\widehat{\pi}$ in non-nominal 
  conditions under adversarial perturbations~$\delta$ on the state
  measurement. Panel (b) illustrates the Lipschitz-constrained
  policy learning scheme implemented on the expert generated~dataset
  to obtain policy~$\widehat{\pi}$.}
  \label{fig: pipeline}
\end{figure}

Our objective is to learn a feedback policy 
from the dataset $X, U$ of expert demonstrations to solve the control
task~\eqref{eq:opt_control_value_function} while remaining robust to
(adversarial) disturbances~$\delta$ of full-state measurements. 
To this end, we seek an optimization-based learning formulation 
that allows us to explicitly constrain the sensitivity of the learned 
policy to (adversarial) disturbances. The Lipschitz constant of the
learned policy serves as a measure of its sensitivity to disturbances,
and we thereby formulate the (adversarially) robust policy learning problem
as a Lipschitz-constrained policy learning problem~\cite{VK-AAALM-FP:20}:
\begin{align}\label{eq:Opt_prob}
  \begin{aligned}
     &\underset{\pi \in \Lip(B_r(0); \real^m)}{\min} 
     					\quad  \frac{1}{N T}\sum_{i=1}^N \sum_{t=0}^{T-1} L \left(\pi ( x_t^{(i)} ), u_t^{(i)} \right), \\
  &\qquad\quad\text{s.t.} \qquad\quad\; \lip(\pi) \leq \alpha,
  \end{aligned}
\end{align}
where $L$ is a strictly convex and Lipschitz continuous loss function 
for the learning problem, $\lip(\pi)$ is
the Lipschitz constant of the policy $\pi$, and
$\alpha\in \mathbb{R}_{\geq 0}$ is a target upper bound for the
Lipschitz constant of the learned policy $\widehat{\pi}$ (the
minimizer in~\eqref{eq:Opt_prob}). The Lipschitz constraint
in~\eqref{eq:Opt_prob} serves as a mechanism to induce robustness of
the learned policy to disturbances~$\delta$ (the smaller the parameter $\alpha$, 
the more robust the policy $\widehat{\pi}$ is
to the disturbances $\delta$~\cite{MF-AR-HH-MM-GJP:19}). 
Figure~\ref{fig: pipeline} illustrates our setup.

\medskip
We then use the robustness of the learned policy, along with
a bound on the training loss, to obtain guarantees on the
closed-loop performance under the learned policy~$\widehat{\pi}$.
For this, we first combine the training loss with the Lipschitz bound
to obtain a bound on the worst-case error 
$\left\| \widehat{\pi} - \pi^* \right\|_\infty$, where $\left\| \widehat{\pi} - \pi^* \right\|_\infty
= \sup_{x \in B_r(0)} \left\| \widehat{\pi}(x) - \pi^*(x) \right\|$.
Then for a given worst-case learning error bound 
$\left\| \widehat{\pi} - \pi^* \right\|_\infty \leq \varepsilon$,
we obtain via our analysis
(i) a robust closed-loop stability bound as a function of~$\varepsilon$, and
(ii) bounds on closed-loop performance (measured in terms of the cost~\eqref{eq: nonlinear system task} 
incurred on the task) as a function of~$\varepsilon$.
Conversely, in order to satisfy target bounds on stability and performance, 
our analysis can be used to obtain a target bound on~$\varepsilon$ 
which must be satisfied by the learned policy, if
some additional information on the system and task are available.

We now develop appropriate notions of closed-loop performance 
and robustness under the feedback policies learned from
expert demonstrations.  We note that the control
task~\eqref{eq:opt_control_value_function}, being one of optimal
control of System~\eqref{eq: dynamical model}, has a natural
performance metric given by the value function.  Let
$V^{\widehat{\pi}}$ be the value function associated with the learned
feedback policy~$\widehat{\pi}$ for~System~\eqref{eq: dynamical
  model}:
\begin{align} \label{eq:inf_hor_cost_feedback}
\begin{aligned}
	V^{\widehat{\pi}}(x) = \sum_{t=0}^\infty \gamma^t c_{\pi} \left( f_{{\widehat{\pi}}}^t(x) \right).
\end{aligned}
\end{align} 
Since the expert implements the optimal policy~$\pi^*$, the
performance of the learned policy can be measured by its regret
with respect to the expert policy~$\pi^*$.  The regret associated with the
learned policy~$\widehat{\pi}$ relative to the expert policy~$\pi^*$ is:
\begin{align}\label{eq: regret}
  {\mc R}(\widehat{\pi}) = \sup_{x \in B_{r}(0)}\left\{ V^{\widehat{\pi}}(x) - V^*(x) \right\}.
\end{align}
When $\mathcal{R}(\widehat{\pi}) = 0$, the
performance of the learned policy equals the performance of an optimal
policy for the control task \eqref{eq:opt_control_value_function}. 
Conversely, the performance of the learned policy degrades 
as $\mathcal{R}(\widehat{\pi})$ increases.
Naturally, the objective of the policy learning problem
is now to minimize the regret incurred by the learned policy~$\widehat{\pi}$.
Note that this is a more important performance metric in the closed-loop
setting than the loss function~$L$ used for learning in~\eqref{eq:Opt_prob},
as it encodes the cost incurred by the evolution of the 
system under the learned feedback policy.
We now note that the regret~$\mc R$ only measures
the performance of the learned policy under nominal conditions
(in the absence of perturbations on the state measurements)
and does not shed light on its performance in the presence of
adversarial perturbations. This calls for an appropriate robustness
metric, for which we will use the regret associated with the 
policy~$\widehat{\pi}$ when subject to perturbations relative 
to when deployed under nominal conditions, that is,
\begin{align}\label{eq: robust regret}
  {\mc S}(\widehat{\pi}) = \sup_{x \in B_{r}(0)}\left\{ V^{\widehat{\pi}_\delta}(x) - V^{\widehat{\pi}}(x) \right\},
\end{align}
where $\widehat{\pi}_\delta(x) = \widehat{\pi} (x+\delta)$.
Intuitively, if ${\mc S}(\widehat{\pi})$ is small, then the
performance of the policy $\widehat{\pi}$ under perturbation is close
to its performance in nominal conditions, and $\widehat{\pi}$ is
robust to feedback perturbations. Again, we note that this robustness
metric measures closed-loop robustness by encoding the cost incurred
by the evolution of the system under the learned feedback policy
subject to feedback perturbations.  We would ideally like to keep both
$\mc R$ and $\mc S$ low, which would imply that the policy performs
well both under nominal conditions and when subjected to feedback
perturbations.  However, we shall see later that there may exist
tradeoffs between the two objectives, presenting an
obstacle to such a~goal.

We now address some crucial technical issues arising in the closed-loop dynamic setting 
in relation to minimizing the performance metrics~$\mc R$ and~$\mc S$. 
We note that the policy learning problem~\eqref{eq:Opt_prob}
is formulated over the set $B_r(0) \in \real^n$, which is the region of 
interest containing the data from expert demonstrations.
Now, in order to measure the performance
of a learned policy~$\widehat{\pi}$ using the metrics~$\mc R$ and~$\mc S$,
we must first ensure that the closed-loop trajectories of the system,
under policy~$\widehat{\pi}$, remain in $B_r(0)$ (for initial
conditions in $B_r(0)$). In the absence of such a guarantee,
the metrics~$\mc R$ and~$\mc S$ are likely to be
unbounded, and would therefore not serve as useful 
measures of performance. We therefore obtain 
robust stability bounds that specify the conditions 
under which closed-loop trajectories remain bounded in~$B_r(0)$.


\section{Robust closed-loop stability and performance}
In this section, we present the theoretical results 
underlying the robust feedback policy learning 
framework outlined in Section~\ref{sec:framework_outline}.
The results are presented in three parts: (i) We begin with an analysis of
the Lipschitz-constrained policy learning problem~\eqref{eq:Opt_prob}.
In Theorem~\ref{thm:finite_sample_bound}, we provide
a finite sample guarantee on the maximum learning error
incurred in the region of interest $B_r(0)$, i.e., 
$\left\| \widehat{\pi} - \pi^* \right\|_\infty$.
This guarantee on the learning error bound is to be combined with
the closed-loop stability and performance guarantees
obtained later for policies satisfying a given learning error bound.
(ii) We then present a closed-loop stability analysis for 
System~\eqref{eq: dynamical model} under learned feedback control 
policies satisfying a given learning error bound.
In Theorem~\ref{thm:stability}-(i) we establish
that the closed-loop system under optimal feedback~$\pi^*$
is exponentially stable. Furthermore, in Theorem~\ref{thm:stability}-(ii)
we establish a robust stability guarantee (to bounded adversarial
disturbances on the state measurements) for learned feedback 
control policies satisfying a given learning error bound.
(iii) We finally present an analysis of performance on 
the control task~\eqref{eq: nonlinear control task} 
under learned feedback control policies 
satisfying a given learning error bound. 
Theorem~\ref{thm:regret_robustness_bound}-(i) provides an upper bound on
the regret incurred by the learned policy with respect to the expert policy. 
Theorem~\ref{thm:regret_robustness_bound}-(ii) quantifies the robustness of
the closed-loop performance in terms of the Lipschitz constant of the
learned feedback policy.

\subsection{Robust stability with learned feedback policy}
We first present the following result on the quadratic boundedness
of the optimal value function:
\begin{lemma}[\bf \emph{Quadratic boundedness of optimal value function}] \label{lemma:quad_bound}
There exist $\underline{\kappa}^*, \bar{\kappa}^* \in \real_{\geq 0}$ 
such that $\frac{\mu}{2} \leq \underline{\kappa}^* \leq \bar{\kappa}^*$ 
and the optimal value function in~\eqref{eq:opt_control_value_function} satisfies
$\underline{\kappa}^* \| x \|^2 \leq V^*(x) \leq \bar{\kappa}^* \| x \|^2$. 
\end{lemma}
\medskip
\noindent
We make the following assumption on the constants~$\underline{\kappa}^*, \bar{\kappa}^*$
in Lemma~\ref{lemma:quad_bound}:
\begin{assumption}[\bf \emph{Bounds on~$\underline{\kappa}^*, \bar{\kappa}^*$}]
For~$\gamma \in \left( 1 - \mu/ \left(\lambda \sqrt{1+ {\alpha^*}^2} \right) , 1 \right)$,
the constants~$\underline{\kappa}^*, \bar{\kappa}^*$ in Lemma~\ref{lemma:quad_bound}
are such that $\lambda \sqrt{1+ {\alpha^*}^2} /2 \leq \underline{\kappa}^* \leq \bar{\kappa}^* < \underline{\kappa}^* + \mu/2$.
\end{assumption}
\medskip
\noindent
The following theorem establishes robust stability of the 
closed-loop system under the learned policy~$\widehat{\pi}$
from a bound on the policy error 
$\left\| \widehat{\pi}(x) - \pi^*(x) \right\|_\infty$
and measurement disturbances~$\delta$:
\begin{theorem}[\bf \emph{Robust exponential stability under Lipschitz policy}]
 \label{thm:stability}
  Let $\gamma' = 1 - \mu/(2 \bar{\kappa}^*)$.
  Let~$\pi^*$ be the minimizer in~\eqref{eq:opt_control_value_function} 
  for some $\gamma \in \left( \bar{\kappa}^* \gamma'/ \underline{\kappa}^* \; , \; 1 \right)$,
  and let~$\widehat{\pi}$ be any policy such that 
  $\left\| \widehat{\pi} - \pi^* \right\|_\infty \leq \varepsilon$
  and $\lip \left( \widehat{\pi} \right) \leq \alpha$.
%
%
Let $\alpha \zeta + \varepsilon \leq \frac{1}{\ell_{f}^u} \left( 1 - \sqrt{\frac{\bar{\kappa}^* \gamma'}{\underline{\kappa}^* \gamma}} \right) r$
and let $\left\| \delta_t \right\| \leq \zeta$ for all $t \in \mathbb{N}$. 
For the closed-loop trajectory $f_{\widehat{\pi}_\delta}^t(x)$
starting from~$x \in B_{r}(0)$ and generated by
the policy $\widehat{\pi}_\delta$, the following holds:
\begin{align*}
\small
\begin{aligned}
	&\left\| f_{\widehat{\pi}_\delta}^t(x) - f_{\pi^*}^t(x) \right\|\\
	& \leq 
	\min \left \lbrace	 \Delta_t , ~2 \left[ \frac{\bar{\kappa}^* \gamma'}{\underline{\kappa}^* \gamma} \right]^{\frac{t}{2}} \| x \|
						              + \ell_{f}^u  \left[ \frac{1 - \left[ \frac{\bar{\kappa}^* \gamma'}{\underline{\kappa}^* \gamma} \right]^{\frac{t}{2}} }{1 - \left[ \frac{\bar{\kappa}^* \gamma'}{\underline{\kappa}^* \gamma} \right]^{\frac{1}{2}}} \right] (\alpha \zeta + \varepsilon), ~2r  \right \rbrace
\end{aligned}
\end{align*}
where $\Delta_t = \begin{cases} \left[ \frac{1 - \ell_{f_{\pi^*}}^t}{1- \ell_{f_{\pi^*}}} \right] \cdot \ell_f^u (\alpha \zeta + \varepsilon), \quad ~&\text{if}~ \ell_{f_{\pi^*}} \neq 1, \\ \smallskip
													t \cdot \ell_f^u (\alpha \zeta + \varepsilon), \quad ~&\text{if}~ \ell_{f_{\pi^*}} = 1.  \end{cases}$
\end{theorem}
\medskip
\noindent
We refer the reader to Appendix~\ref{app:robust_stability} for the proof.
The robust stability result can be understood in the sense of input-to-state
stability~\cite{ZPJ-EDS-YW:01,EDS:08}, in that we exploit the
exponential stability result for the expert policy~$\pi^*$ and treat
the learned policy~$\widehat{\pi}$ as a perturbation on~$\pi^*$.  By
obtaining boundedness of the learning error along the closed-loop
trajectory, we establish that the closed-loop trajectory under the
learned policy both stays within a bounded region around the optimal
trajectory and converges asymptotically to a bounded region around
the~origin.

\subsection{Regret and robustness with learned feedback policy}
Having clarified the issue of robust stability,
we now present a regret analysis for the learned control policy
$\widehat{\pi}$. 
We first present the following lemma on an incremental
exponential stability property of exponentially stabilizing
Lipschitz feedback policies on~$B_r(0)$:
\begin{lemma}[\bf \emph{Incremental exponential stability}] \label{lemma:inc_exp_stability}
Let~$\pi$ be an exponentially stabilizing Lipschitz feedback 
policy for System~\eqref{eq: dynamical model}
such that~$\left\| f_\pi^t(x) \right\| \leq M \beta^t \| x \|$ for some
$M \in \real_{\geq 0}$ and $\beta \in (0,1)$ and $B_r(0)$ is $f_\pi$-invariant. 
For $x, x' \in B_r(0)$, there exists $M(x_1, x_2) \in \real_{\geq 0}$
such that $\left\| f_\pi^t(x) - f_\pi^t(x') \right\| \leq M(x_1,x_2) \beta^t \| x_1 - x_2 \|$.
\end{lemma}
\medskip
\noindent
We make the following assumption on the existence of a 
uniform bound on~$M(x_1, x_2)$ in Lemma~\ref{lemma:inc_exp_stability} 
over $x_1, x_2 \in B_r(0)$:
\begin{assumption}[\bf \emph{Uniform incremental exponential stability}] \label{ass:uniform_inc_stability}
The optimal policy~$\pi^*$ in~\eqref{eq:opt_control_value_function}
is such that for any~$\pi \in \Lip(B_r(0); \real^m)$ satisfying 
$\left\| \pi - \pi^* \right\|_\infty \leq \left( 1 - \sqrt{\bar{\kappa}^* \gamma'/(\underline{\kappa}^* \gamma)} \right) r / \ell_{f}^u$,
there exists a constant~$M = \sup_{x_1, x_2 \in B_r(0)} M(x_1,x_2)$
in Lemma~\ref{lemma:inc_exp_stability}.
\end{assumption}
\medskip
\noindent
The following theorem establishes a bound on the sub-optimality of the
closed-loop performance of system~\eqref{eq: dynamical model} with
$\widehat{\pi}$ and a robustness bound for the deterioration of the 
closed-loop performance under bounded disturbances:
\begin{theorem}[\bf \emph{Regret and robustness of learned policy}] \label{thm:regret_robustness_bound}
  Let $\gamma'$ be as specified in Theorem~\ref{thm:stability},
  $M, \beta$ be as in Lemma~\ref{lemma:inc_exp_stability} and Assumption~\ref{ass:uniform_inc_stability},
  and let~$\pi^*$ be the minimizer in~\eqref{eq:opt_control_value_function} 
  for some $\gamma \in \left( \bar{\kappa}^* \gamma'/ \underline{\kappa}^* \; , \; 1 \right)$.
  Let $\varepsilon \leq \left( 1 - \sqrt{\bar{\kappa}^* \gamma'/(\underline{\kappa}^* \gamma)} \right) r / \ell_{f}^u$,
  $\widehat{\pi}$ be any policy such that $\left\| \widehat{\pi} - \pi^* \right\|_\infty \leq \varepsilon$
  and $\lip \left( \widehat{\pi} \right) \leq \alpha$.
  Furthermore, let~$\Theta = M^2/(1 - \gamma \beta^2)$. \\\\
 \noindent
  {\bf \emph{(i) Regret:}} The regret~$\mathcal{R}$ of the policy~$\widehat{\pi}$
  relative to~$\pi^*$, as defined in~\eqref{eq: regret}, satisfies:
  \begin{align*}
    {\mc R}(\widehat{\pi}) 
    \leq 
    \frac{\lambda}{1 - \gamma}	\left[ c_1 r \sqrt{1 + \left| \max \left \{ \alpha , \alpha^* \right \} \right|^2 } ~\Delta+ \frac{1}{2} c_2 \Delta^2 \right],
  \end{align*}
where 
\begin{align*}
\begin{aligned}
	c_1 &=  1   + \gamma \Theta \ell_{f}^u ,
\qquad
c_2 = 1 + \gamma \Theta \sqrt{1+ {\alpha^*}^2}~  {(\ell_{f}^u)}^2,\\
\Delta&=\left\| \widehat{\pi} - \pi^* \right\|_{\infty}.
\end{aligned}
\end{align*}
\noindent
  {\bf \emph{(ii) Robustness:}} Let $\| \delta_t \| \le \zeta$ for any~$t \in \mathbb{N}$. 
  For any $\gamma \in (0, 1)$, the robustness metric~$\mathcal{S}$ of the 
  policy~$\widehat{\pi}$, as defined in~\eqref{eq: robust regret}, satisfies:
  \begin{align*}
   {\mc S}(\widehat{\pi}) 	
   \leq	
   \frac{\lambda \alpha}{1 - \gamma}	\left[ d_1 r \alpha \sqrt{1 + \alpha^2 } ~\zeta + \frac{1}{2} d_2 \alpha^2 \zeta^2 \right],
  \end{align*}
  where 
\begin{align*}
\begin{aligned}
	d_1 =  1   + \gamma \Theta  \ell_{f}^u ,
\qquad	d_2 = 1 + \gamma \Theta \sqrt{1+ {\alpha}^2}~  {(\ell_{f}^u)}^2 .
\end{aligned}
\end{align*}
\end{theorem}

\medskip 
\noindent
Theorem~\ref{thm:regret_robustness_bound}-(i)
establishes that the regret bound for the learned policy scales
linearly with the deviation of the learned policy from the expert (optimal) policy.
We also note that the regret bound scales with~$\lambda$, the
Lipschitz constant of the gradient of the stage cost, and the Lipschitz
constant of the dynamics (w.r.t.~$u$), 
as they modulate the sensitivity to variations of the input.
 Furthermore, we want the performance of the learned policy under
 disturbances to be close its nominal performance, i.e., 
 a low value of~$\mathcal{S}$.
 Theorem~\ref{thm:regret_robustness_bound}-(ii)
 establishes that the robustness of performance is determined
 by the sensitivity of the learned policy to disturbances,
 in particular that the robustness bound scales linearly 
 with the Lipschitz constant of the learned policy.
Theorem~\ref{thm:regret_robustness_bound}-(ii)
provides the designer with a robustness guarantee while implementing 
the learned policy in the presence of bounded (possibly adversarial) disturbances
to measurements.

  Furthermore, we note that in the limit $N \rightarrow \infty$ of the size of the dataset, 
  Theorem \ref{thm:regret_robustness_bound} suggests a tradeoff between the regret
  ${\mc R}(\widehat{\pi})$ and the robustness metric ${\mc S}(\widehat{\pi})$ as we vary the Lipschitz
  bound~$\alpha$ in \eqref{eq:Opt_prob}. As we decrease $\alpha$,
  the deviation of the learned policy $\widehat{\pi}$ from the optimal
  policy $\pi^*$ increases, 
  and so does the bound in Theorem~\ref{thm:regret_robustness_bound}-(i)
  (via an increase in~$\varepsilon$). 
  Instead, as we increase~$\alpha$ such that the constraint in~\eqref{eq:Opt_prob} 
  is no longer active, the
  learned policy converges to the optimal policy $\pi^*$, and the
  bound in Theorem~\ref{thm:regret_robustness_bound}-(i) decreases to
  zero. Similarly, as we decrease $\alpha$, the Lipschitz constant of
  the learned policy, $\ell_{\widehat{\pi}}$, decreases, and so does
  the bound in Theorem~\ref{thm:regret_robustness_bound}-(ii). See Fig. \ref{fig:
    tradeoff} in Section~\ref{sec:numerical_resuls} for an
  illustration of this tradeoff. 
Furthermore, we see that strong convexity of the cost induces stability
properties and $\lambda$-smoothness allows for the tuning of regret.

\section{Lipschitz-constrained policy learning}
We now present results from our analysis of the 
Lipschitz constrained policy learning problem~\eqref{eq:Opt_prob}.
We note that the training data for the feedback policy learning 
problem~\eqref{eq:Opt_prob} consists of evaluations of the 
expert policy~$\pi^*$ over a finite set of points $\lbrace x_t^{(i)} \rbrace \subset B_r(0)$ 
in the state space, and the objective is to generalize over the region of interest $B_r(0)$.
The following theorem establishes a (maximum) generalization error bound for the 
minimizer~$\widehat{\pi}$ over the region~$B_r(0)$:
\begin{theorem}[\bf \emph{Finite sample guarantees on Lipschitz policy learning}] \label{thm:finite_sample_bound}
Let $\widehat{\pi}$ be a minimizer in~\eqref{eq:Opt_prob}.
For any $\delta > 0$, the maximum learning error in $B_r(0)$ satisfies:
\begin{align*}
	&\mathbb{P} \bigg[ \left\| \widehat{\pi} - \pi^* \right\|_\infty > (\alpha + \alpha^*)  \delta  + \varepsilon_{\rm train} (\widehat{\pi})  \bigg] \\
	& \qquad < \sum_{k=1}^{\left| \mathcal{E}_{\delta/2} \right|} (-1)^{k+1} {\left| \mathcal{E}_{\delta/2} \right| \choose k} \left[ 1 - \frac{k \cdot c(k)}{\mathrm{Vol} \left( B_r(0) \right) } \right]^{N},
\end{align*}
\medskip
\noindent
where:
\begin{align*}
	c(k) &= 
		\begin{cases} 
				\mathrm{Vol} \left( B_r(0) \cap B_{\delta/2}(z \in \partial B_r(0)) \right), &\quad k ~\text{is~odd} \\ 
				\mathrm{Vol} \left( B_r(0) \cap B_{\delta / 2 \nu^T} (0) \right) , &\quad k ~\text{is~even} 
		\end{cases} \\ 
	\left| \mathcal{E}_{\delta/2} \right| &\leq \frac{\mathrm{Vol} \left( B_r(0) \right)}{\mathrm{Vol} \left( B_r(0) \cap B_{\delta/4}(z \in \partial B_r(0) ) \right)}, \\
	\nu &= \sqrt{\frac{\underline{\kappa}^* - \lambda \sqrt{1 + {\alpha^*}^2}/2}{\gamma \bar{\kappa}^*}}.
\end{align*}
\end{theorem}
\medskip
\noindent
We refer the reader to Appendix~\ref{app:finite_sample_bound_proof} for a proof
of this result. Theorem~\ref{thm:finite_sample_bound} 
shows that although a larger~$\alpha$ allows for achieving a lower~$\varepsilon_{\rm train}$,
it can result in worse generalization performance. This is due to the fact that  
the $(\alpha + \alpha^*) \delta$ term in the bound scales linearly with~$\alpha$,
which can potentially result in a higher maximum learning error 
$\left\| \widehat{\pi} - \pi^* \right\|_\infty$.
Furthermore, from Appendix~\ref{app:finite_sample_bound_proof}-(c)
in the proof of Theorem~\ref{thm:finite_sample_bound}, 
we remark that the probabilistic bound in Theorem~\ref{thm:finite_sample_bound} 
is a worst-case bound which can potentially be tightened. In particular,
the tightness of the estimate provided by the bound worsens with 
an increase in the length~$T$ of the control horizon in the demonstrations.

We finally note that Theorems~\ref{thm:stability} and~\ref{thm:regret_robustness_bound}
establish robust stability and performance bounds for policies~$\widehat{\pi}$ that
satisfy (i) $\| \widehat{\pi} - \pi^* \|_\infty \leq \varepsilon$,
and (ii) $\lip(\widehat{\pi}) \leq \alpha$,
whereas Theorem~\ref{thm:finite_sample_bound} yields a
(probabilistic) bound on the violation of the condition 
$\| \widehat{\pi} - \pi^* \|_\infty \leq \varepsilon$
for finite datasets of size~$N$ (while the Lipschitz bound still holds).
Therefore, by combining the bounds in Theorems~\ref{thm:stability} and~\ref{thm:regret_robustness_bound}
with the finite sample bound in Theorem~\ref{thm:finite_sample_bound},
we obtain the desired closed-loop generalization and robustness bounds.


%
\begin{algorithm}[!b]
\caption{Graph-based Lipschitz policy learning}
\textbf{Input:} Training data, Graph size~$n$, Number of edges~$|\mc{E}|$, Lipschitz bound~$\alpha$, Number of iterations~$k$ 
\begin{algorithmic}[1]
	\State Sample~$n$ points (graph vertices) uniformly i.i.d. from~$B_r(0)$
	\State Partition training dataset as in~\eqref{eq:dataset_partition} 
	\State Implement~$k$ iterations of primal-dual algorithm~\eqref{eq:p-d_Lipschitz_const_loss_min}
\end{algorithmic}
\textbf{Output:} Minimizer $\widehat{\mathbf{u}}$ 
	\label{alg:lip_policy_learning}
\end{algorithm}
We now present a graph-based Lipschitz policy learning algorithm to
solve~\eqref{eq:Opt_prob}.
We sample~$n$ points $\left \lbrace X_i \right \rbrace_{i=1}^n$,
uniformly i.i.d from $B_r(0)$. Considering the points
$\left \lbrace X_i \right \rbrace_{i=1}^n$ as the (embedding of)
vertices, we construct an undirected, weighted, connected
graph~$\mathcal{G} = (\mathcal{V}, \mathcal{E} )$, with vertex
set~$\mathcal{V} = \lbrace 1, \ldots, n \rbrace$, edge
set~$\mathcal{E} \subseteq \mathcal{V} \times \mathcal{V}$.
We then define a partition~$\mathcal{W} = \lbrace \mathcal{W}_i \rbrace_{i=1}^n$ of the
training dataset~$D = \lbrace x_t^{(i)} \rbrace$ (set of points in the state space
where evaluations of the expert policy are available) as follows:
\begin{align}\label{eq:dataset_partition}
  \mathcal{W}_i = \left \lbrace  x \in D \; | \;  \left| x - X_i
  \right| \leq \left| x - X_j \right| ~\forall~ j \in \mathcal{V}
  \setminus \lbrace i \rbrace  \right \rbrace .
\end{align}
Finally, we write the discrete (empirical) Lipschitz-constrained policy learning 
problem over the graph~$\mathcal{G}$ as follows 
(which can be viewed as the discretization 
of~\eqref{eq:Opt_prob} over the graph~$\mathcal{G}$):
\begin{align}\label{eq:empirical_loss_min_graph_discretization}
\begin{aligned}
  &\min_{ \substack{\widehat{\mathbf{u}} = (\widehat{u}_1, \ldots, \widehat{u}_n), \\ \widehat{u}_i \in \real^m } }
  ~ \sum_{i \in \mathcal{V}} \sum_{j \in \mathcal{W}_i} L ( \widehat{u}_i , u_j ), \\ 
 &\qquad \; \text{s.t.}~~ \left| \widehat{u}_r - \widehat{u}_s \right| \leq \alpha \left| X_r - X_s \right|, ~~\forall~(r,s) \in \mathcal{E}.
\end{aligned}
\end{align}
We note that Problem~\eqref{eq:empirical_loss_min_graph_discretization} is convex
(strictly convex objective function with convex constraints) and the
corresponding Lagrangian is given by:
\begin{align*}
\begin{aligned}
	\mathcal{L}_{\mathcal{G}}(\widehat{\mathbf{u}}, \Lambda) =& \sum_{i \in \mathcal{V}} \Bigg[ \sum_{s \in \mathcal{W}_i} L ( \widehat{u}_i, u_s )\\ 
	&+  \frac{1}{2} \sum_{j \in \mathcal{V}} \lambda_{ij} \left( \left| \widehat{u}_i - \widehat{u}_j \right|^2 - \alpha \left| X_i - X_j \right|^2 \right) \Bigg],
\end{aligned}
\end{align*}
where~$\Lambda = [\lambda_{ij}]_{i,j=1}^n$ is the matrix of Lagrange
multiplier for the pairwise Lipschitz constraints. Define a
primal-dual dynamics for the
Lagrangian~$\mathcal{L}_{\mathcal{G}}(\widehat{\mathbf{u}}, \Lambda)$ with
time-step sequence~$\lbrace h(k) \rbrace_{k \in \mathbb{N}}$:
\begin{align}\label{eq:p-d_Lipschitz_const_loss_min}
\small
\begin{aligned}
	\widehat{\mathbf{u}}(k+1) &= \widehat{\mathbf{u}}(k) - h(k) ~\nabla_{\widehat{\mathbf{u}}} \mathcal{L}_{\mathcal{G}} \left( \widehat{\mathbf{u}}(k)  ,  \Lambda(k) \right), \\
	 \Lambda(k+1) &= \max \lbrace 0\; , \; \Lambda(k) + h(k) ~\nabla_{\Lambda} \mathcal{L}_{\mathcal{G}} \left(\widehat{\mathbf{u}}(k)  ,  \Lambda(k) \right) \rbrace.
\end{aligned}
\end{align}
The primal dynamics is a discretized heat flow over the
graph~$\mathcal{G}$ with a weighted Laplacian,
where~$\nabla_{\widehat{\mathbf{u}}} \mathcal{L}_{\mathcal{G}} \left(
  \widehat{\mathbf{u}}(k) , \Lambda(k) \right) = \left( \Delta(\Lambda)
  \otimes I_{\dim(\mathbb{Y})} \right) \widehat{\mathbf{u}} 
  + \nabla_1 L (\widehat{\mathbf{u}}, \mathbf{u})$, and~$\Delta(\Lambda)$ is
the $\Lambda$-weighted Laplacian of the graph~$\mathcal{G}$.  
The convergence of the
solution~$\lbrace (\widehat{\mathbf{u}}(k), \Lambda(k)) \rbrace_{k \in
  \mathbb{N}}$ of the primal-dual
dynamics~\eqref{eq:p-d_Lipschitz_const_loss_min} to the saddle point
of the Lagrangian~$\mathcal{L}_{\mathcal{G}}$
follows~\cite{KA-HA-LH-HU:58} from the convexity of
Problem~\eqref{eq:empirical_loss_min_graph_discretization}.

\begin{figure}[!t]
  \centering
  \includegraphics[width=0.75\columnwidth]{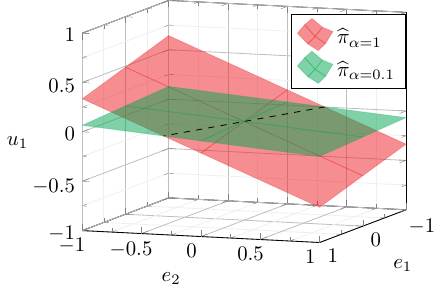}
  \caption{This figure shows the surface of policy $\widehat{\pi}$ in the state space for system \eqref{eq: error dynamics}, which is learned using Alg. \ref{alg:lip_policy_learning} with $\alpha=1$ (red surface) and $\alpha=0.1$ (green surface), and the expert being the LQR for system \eqref{eq: error dynamics}.}
    \label{fig: learned_policy}
\end{figure}
\begin{figure}[!t]
  \centering
  \includegraphics[width=1\columnwidth,trim={0cm 0cm 0cm 0cm},clip]{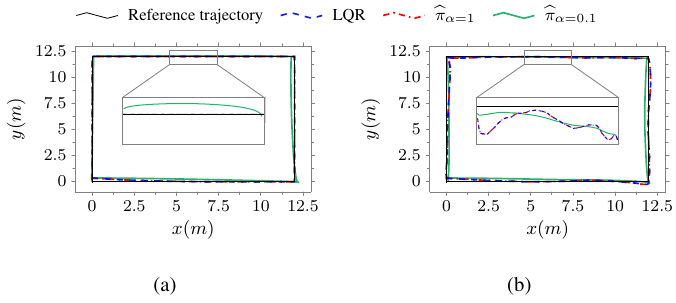}
  \caption{Panel (a) and panel (b) show the trajectory tracking performance for the LQR (dashed blue line), the learned policy $\widehat{\pi}$ learned using Alg. \ref{alg:lip_policy_learning} with $\alpha=1$ (dash-dotted red line) and $\alpha=0.1$ (solid green line). In panel (a), the policies are deployed in nominal conditions. The policy $\widehat{\pi}_{\alpha=1}$ performs as good as the LQR while the policy $\widehat{\pi}_{\alpha=0.1}$ performs poorly compared to the LQR and $\widehat{\pi}_{\alpha=1}$. In panel (b), the policies are deployed in non-nominal conditions. The performance of the LQR and policy $\widehat{\pi}_{\alpha=1}$ is worse than when deployed in nominal conditions, while the performance of policy $\widehat{\pi}_{\alpha=0.1}$ in non-nominal conditions remains almost the same as in nominal conditions.}
  \label{fig: traj_tracking}
\end{figure}
\section{Numerical experiments}\label{sec:numerical_resuls}
In this section, we present the results from numerical experiments
applying our algorithm to (i) learn the Linear Quadratic Regulator (LQR), 
and (ii) learn nonlinear control for a nonholonomic system (differential drive mobile robot).
\subsection{Learning the Linear Quadratic Regulator}\label{subsec: LQR}
 We consider a vehicle obeying the following dynamics (see also
\cite{AAALM-VK-FP:19b} and \cite{AAALM-VK-FP:19a}):
\begin{align}\label{eq: car's dynamics}
\small
\begin{split}
  x_{t+1} \!= \!
  \underbrace{
  \begin{bmatrix}
    1 & T_{\text{s}} & 0 & 0\\
    0 & 1 & 0 & 0\\
    0 & 0& 1 & T_{\text{s}}\\
    0 & 0 & 0 & 1
  \end{bmatrix}}_A
                x_t \!+\! 
                \underbrace{
                \begin{bmatrix}
                  0 & 0\\
                  T_{\text{s}} & 0\\
                  0 & 0\\
                  0 & T_{\text{s}}
                \end{bmatrix}}_B
                      u_t, \quad
y_t =x_t + \delta_t
\end{split}
\end{align}
where $x_t \in \mathbb{R}^4$ contains the vehicle's position and
velocity in cartesian coordinates, $u_t \in \mathbb{R}^2$ the input
signal, $y\in\mathbb{R}^4$ the state measurement,
$\delta_t \in \mathbb{R}^4$ bounded measurement noise with
$\|\delta_t\|\leq \zeta$ and $\zeta \in \mathbb{R}_{\geq 0}$, and
$T_{\text{s}}$ the sampling time. We consider the problem of tracking
a reference trajectory, and we write the error dynamics and the
controller as
\begin{align}
e_{t+1}=Ae_t+B\bar{u}_t, \qquad
u_t=\underbrace{-K(e_t+\delta_t)}_{\bar{u}_t} + v_t,
\label{eq: error dynamics}
\end{align}
where $e_t = x_t-\xi_t$ is the error between the system state and the
reference state, $\xi_t \in \mathbb{R}^4$ at time $t$,
$v_t \in \mathbb{R}^2$ is the control input generating $\xi_t$, and
$K$ denotes the control gain. We consider the expert policy to
correspond to the optimal LQR gain, $K_{\text{lqr}}$, which minimizes
a discounted value function as in
\eqref{eq:opt_control_value_function} but with horizon $T$, quadratic
stage cost
$\left. c(e_t,\bar{u}_t) =
  e_t^{\transpose}Qe_t+\bar{u}_t^{\transpose}R\bar{u}_t \right.$ with
error and input weighing matrices $Q\succeq 0$ and $R \succ 0$,
respectively.
Notice that the quadratic stage cost is strongly convex and Lipschitz
bounded over bounded space $e \in B_r(0) \subset \mathbb{R}^4$ and
$\bar{u} \in \mathbb{R}^2$.\\
\textbf{Expert demonstrations.} We generate $N$ expert trajectories using
\eqref{eq: error dynamics} with $K=K_{\text{lqr}}$, $T_{\text{s}}=0.1$, $\gamma=0.82$,
$Q=0.1I_4$, $R=0.1I_2$, and
$\delta_t=0$, contained in the data matrices
$E, U$:
\begin{align*}
E =\left[ \begin{matrix} \mathbf{e}^{(1)} & \ldots & \mathbf{e}^{(N)} \end{matrix} \right], 
		\quad  U =\left[ \begin{matrix} \mathbf{u}^{(1)} & \ldots & \mathbf{u}^{(N)} \end{matrix} \right],
\end{align*}
with $\mathbf{e}^{(i)} = ( e_0^{(i)}, \ldots, e_T^{(i)} )$ and
$\mathbf{u}^{(i)} = ( u_0^{(i)}, \ldots, u_{T-1}^{(i)} )$. Each
trajectory is generated with random initial condition,
$e_0^{ ( i )}\in B_2(0)$ for $i=1,\dots,N$. Note that, since the initial conditions, $e_0^{(i)}$, are inside $B_2(0)$ and $K=K_{\text{lqr}}$ is stabilizing, then, all the data points in $E$ are inside $B_2(0)$.\\
\begin{figure}[!t]
  \centering
  \includegraphics[width=1\columnwidth]{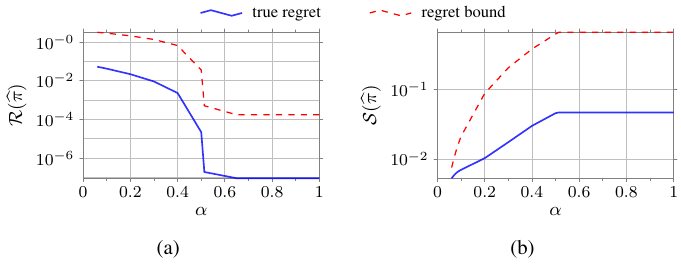}
  \caption{Panel (a) and panel (b) show the true regrets
    $R(\widehat{\pi})$ and $S(\widehat{\pi})$ in \eqref{eq: regret}
    and \eqref{eq: robust regret} (solid blue line), and the regret
    bounds in Theorem~\ref{thm:regret_robustness_bound} 
    (dashed red line) as a function of the
    Lipschitz bound, $\alpha$, in \eqref{eq:Opt_prob},
    respectively. The regret $R(\widehat{\pi})$ and the bound in
    Theorem~\ref{thm:regret_robustness_bound}-(i) decrease as $\alpha$ increases, as
    shown in panel (a), while The regret $S(\widehat{\pi})$ the bound
    in Theorem~\ref{thm:regret_robustness_bound}-(ii) increase with $\alpha$, as
    shown in panel (b).}
  \label{fig: regret_bounds}
\end{figure}\\
\textbf{Policy learning.} Using Alg. \ref{alg:lip_policy_learning}, we learn policy $\widehat{\pi}$ with $\alpha=1$ and $\alpha=0.1$ denoted by $\widehat{\pi}_{\alpha=1}$ and $\widehat{\pi}_{\alpha=0.1}$, respectively. Fig. \ref{fig: learned_policy} shows the surface of the learned policies $\widehat{\pi}_{\alpha=1}$ and $\widehat{\pi}_{\alpha=0.1}$ in the state space. Note that, since the Lipschitz constant of the expert policy, $\pi^*=K_{\text{lqr}}$, is $\ell_{\pi^*}=\|K_{\text{lqr}}\|_2=0.51<\alpha=1$, we get $\| \widehat{\pi}_{\alpha=1} - \pi^*\|_2=0$, which implies that $\widehat{\pi}_{\alpha=1} $ learns exactly the expert policy. On the other hand, since $\alpha=0.1 < \ell_{\pi^*}=0.51$, we get $\| \widehat{\pi}_{\alpha=0.1} - \pi^*\|_2=\epsilon$ for $\epsilon>0$, which implies that $\widehat{\pi}_{\alpha=0.1}$ learns the expert policy with some learning error $\epsilon$. As observed in Fig. \ref{fig: learned_policy}, the Lipschitz constant constraints the slope of the learned surface, where $\widehat{\pi}_{\alpha=0.1}$ has smaller slope than $\widehat{\pi}_{\alpha=1}$, and hence more robust to perturbations in the states. However, smaller Lipschitz constant implies larger learning error, and hence poorer nominal performance.
%
Fig. \ref{fig: traj_tracking} shows the trajectory tracking
performance for the optimal LQR controller, $\widehat{\pi}_{\alpha=1}$, and $\widehat{\pi}_{\alpha=0.1}$. The policies are deployed in nominal conditions, Fig. \ref{fig: traj_tracking}(a), and in non-nominal conditions with $\zeta=0.5$, Fig. \ref{fig: traj_tracking}(b). We observe in Fig. \ref{fig: traj_tracking}(a) that $\widehat{\pi}_{\alpha=1}$ performs better than $\widehat{\pi}_{\alpha=0.1}$ in nominal conditions. On the other hand, we observe in Fig. \ref{fig: traj_tracking}(b) that the performance of $\widehat{\pi}_{\alpha=1}$ degrades when deployed in non-nominal conditions, while the performance of $\widehat{\pi}_{\alpha=0.1}$ remains almost the same, as predicted by
\cite{VK-AAALM-FP:20}.

\begin{figure}[!t]
  \centering
  \includegraphics[width=1\columnwidth,trim={0cm 0.05cm 0cm 0cm},clip]{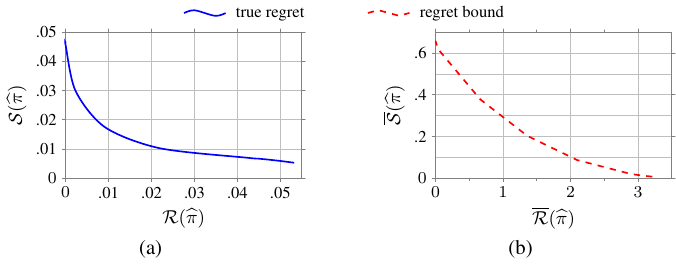}
  \caption{Panel (a) shows the tradeoff between the regrets $R(\widehat{\pi})$ and $S(\widehat{\pi})$ in \eqref{eq: regret} and \eqref{eq: robust regret}, respectively. Panel (b) shows the tradeoff between the upper bounds of $R(\widehat{\pi})$ and $S(\widehat{\pi})$ derived in Theorem~\ref{thm:regret_robustness_bound}, respectively.} 
  \label{fig: tradeoff}
\end{figure}

\textbf{Regret bounds.} The parameters of the bounds in Theorem \ref{thm:regret_robustness_bound} are obtained as follows, $\lambda=\max \{\rho(2Q),\rho(2R)\}$, $\alpha^*=\|K_{\text{lqr}}\|_2$, $\ell_f^u=\|B\|_2$, and $\Theta=\frac{1}{1-\gamma \rho(A+BK)^2}$, where $K$ is a stabilizing gain. Fig. \ref{fig: regret_bounds} shows the regrets ${\mc R}(\widehat{\pi})$ and ${\mc S}(\widehat{\pi})$ in \eqref{eq: regret} and
\eqref{eq: robust regret}, and the corresponding upper bounds derived
in Theorem~\ref{thm:regret_robustness_bound} as a function of the
Lipschitz bound, $\alpha$, in \eqref{eq:Opt_prob}. As can be seen, the
regret ${\mc R}(\widehat{\pi})$ and the corresponding upper bound in Theorem
\ref{thm:regret_robustness_bound}-(i) decrease as $\alpha$ increases, while the
regret ${\mc S}(\widehat{\pi})$ and the corresponding upper bound in 
Theorem~\ref{thm:regret_robustness_bound}-(ii) increase with $\alpha$. Further, the
regrets and the bounds remains constant for $\alpha\geq 0.51$, since
the constraint in \eqref{eq:Opt_prob} becomes inactive and
$\widehat{\pi}$ converges to the optimal LQR
controller. Fig. \ref{fig: tradeoff} shows the tradeoff between the
regrets, as well as the tradeoff between the regrets upper bounds as
we vary the Lipschitz bound, $\alpha$, in \eqref{eq:Opt_prob}. 
This suggests that improving the robustness of the learned policy to
perturbations comes at the expenses of its nominal~performance.

\subsection{Learning nonlinear control for nonholonomic system} \label{subsec: nonholonomic example}
\begin{figure}[!t]
  \centering
  \includegraphics[width=0.5\columnwidth]{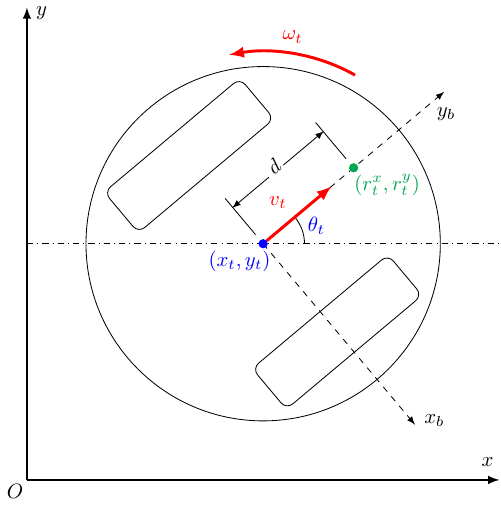}
  \caption{Differential drive mobile robot described in \eqref{eq: nonlinear system}.} 
  \label{fig: nonholonomic}
\end{figure}

\begin{figure}[!t]
  \centering
  \includegraphics[width=0.75\columnwidth]{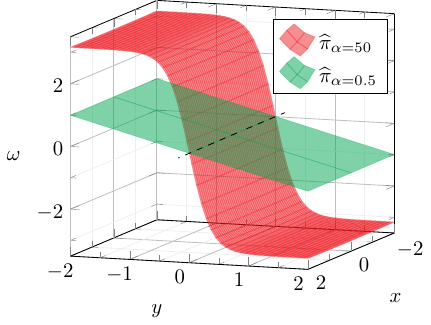}
  \caption{This figure shows the surface of policy $\widehat{\pi}$ that correspond to the input $\omega$ in the subspace $[x,y]^{\T}$ for system \eqref{eq: nonlinear system vector form} for $\theta=0$. Two policies are learned using Alg. \ref{alg:lip_policy_learning} with $\alpha=50$ (red surface) and $\alpha=0.5$ (green surface), and the expert demonstrations are generated as in Appendix \ref{sec: nonholonomic example appendix}.}
    \label{fig: learned_policy_nonlinear}
\end{figure}

\begin{figure}[!t]
  \centering
  \includegraphics[width=1\columnwidth]{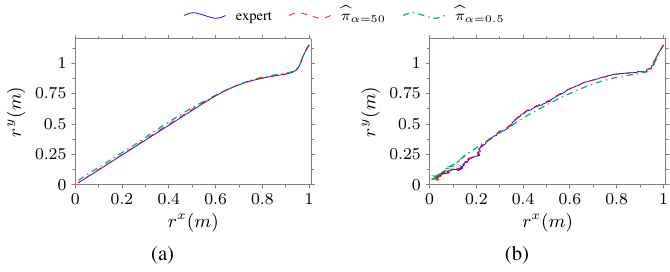}
  \caption{Panel (a) and panel (b) show the trajectory of the expert (solid blue line), the learned policy $\widehat{\pi}$ learned using Alg. \ref{alg:lip_policy_learning} with $\alpha=50$ (dashed red line) and $\alpha=0.5$ (dashdotted green line). In panel (a), the policies are deployed in nominal conditions. The policy $\widehat{\pi}_{\alpha=50}$ outputs the same trajectory as the expert while the policy $\widehat{\pi}_{\alpha=0.5}$ outputs a different trajectory towards the equilibrium. In panel (b), the policies are deployed in non-nominal conditions. The performance of the expert and policy $\widehat{\pi}_{\alpha=50}$ is worse than when deployed in nominal conditions, while the performance of policy $\widehat{\pi}_{\alpha=0.5}$ in non-nominal conditions remains almost the same as in nominal conditions.}
  \label{fig: stabilization}
\end{figure}

We consider nonholonomic differential drive mobile robot (see Fig. \ref{fig: nonholonomic}) obeying the following discrete-time nonlinear dynamics
\begin{align}
\begin{split}\label{eq: nonlinear system}
x_{t+1}&=T_s v_t \cos(\theta_t)+x_t, \qquad \text{for } t\geq 0\\
y_{t+1}&=T_s v_t  \sin(\theta_t)+y_t,\\
\theta_{t+1}&=\theta_t+T_s \omega_t
\end{split}
\end{align}
where $x_t\in \mathbb{R}$ and $y_t \in \mathbb{R}$ are the position of the robot's centroid in the cartesian coordinate frame $(O;x,y)$, $\theta_t \in \mathbb{R}$ is the robot's orientation, $v_t\in \mathbb{R}$ and $\omega_t \in \mathbb{R}$ are the robot's forward and angular velocity at time $t$, respectively, which are the system's inputs, and $T_s >0$ is the sampling time. The dynamics in \eqref{eq: nonlinear system}, can be written in the following vector form
\begin{align}
\begin{split}\label{eq: nonlinear system vector form}
q_{t+1}=f(q_t,u_t),\qquad
u_t= \pi(q_t + \delta_t) \qquad \text{for } t\geq 0,
\end{split}
\end{align}
where $q_t=[x_t, y_t, \theta_t]^{\transpose}$ is the state, $u_t=[v_t, \omega_t]^{\transpose}$ is the input, $f : \mathbb{R}^3 \times \mathbb{R}^2 \rightarrow \mathbb{R}^{3}$ is the dynamics, $\pi : \mathbb{R}^3 \rightarrow \mathbb{R}^{2}$ is the control policy, and $\delta_t \in \mathbb{R}^3$ is a bounded perturbation, with $\|\delta_t\|\leq \zeta \in \mathbb{R}_{\geq 0}$. 
Let $r_t=[r^x_t,r^y_t]^{\transpose}$ be a point fixed on the robot at a fixed distance $d$ from $[x_t,y_t]^{\transpose}$ (see Fig. \ref{fig: nonholonomic}). The task is to stabilize the point $r_t$ at $[0,0]^{\transpose}$, which is described by the following regulator problem
\begin{align}
\begin{split}\label{eq: nonlinear control task}
 &  	\min_{\pi \in \Lip \left( \real^3 ; \real^2 \right)} \; \lim_{T \rightarrow \infty} \sum_{t=0}^{T}\gamma^t \left({r_t}^{\transpose}Qr_t + {u_t}^{\transpose}\mc{R}(\theta_t)^{\transpose}R\mc{R}(\theta_t)u_t\right),\\
 & \qquad \; \text{s.t.}\qquad \begin{cases}
q_{t+1}&=f(q_t,u_t),\\
u_t&= \pi(q_t + \delta_t),
 \end{cases}
 \end{split}\\
&\text{where}\qquad \mc{R}(\theta_t) = \begin{bmatrix}
T_s\cos(\theta_t) & -d T_s \sin(\theta_t)\\
T_s\sin(\theta_t) & d T_s \cos(\theta_t)
\end{bmatrix} \nonumber
\end{align}
where $\gamma$ is the discount factor, and $Q \succeq 0$ and $R \succ 0$ are weighing matrices.\\
\textbf{Expert demonstrations.} We consider the expert policy, $\pi^*$, to be the minimizer of \eqref{eq: nonlinear control task}. The derivation of $\pi^*$ for this example is presented in Appendix \ref{sec: nonholonomic example appendix}.\\
\textbf{Policy learning.} Using Alg. \ref{alg:lip_policy_learning}, we learn policy $\widehat{\pi}$ with $\alpha=50$ and $\alpha=0.5$ denoted by $\widehat{\pi}_{\alpha=50}$ and $\widehat{\pi}_{\alpha=0.5}$, respectively. Fig. \ref{fig: learned_policy_nonlinear} shows the surface of the learned policies $\widehat{\pi}_{\alpha=50}$ and $\widehat{\pi}_{\alpha=0.5}$ that correspond to the input $\omega$ in the subspace $[x,y]^{\T}$ for $\theta=0$. Since the Lipschitz constant of the expert policy, $\pi^*$, is $\ell_{\pi^*}=16.65<\alpha=50$, the policy $\widehat{\pi}_{\alpha=50} $ learns exactly the expert policy. On the other hand, since $\alpha=0.5 < \ell_{\pi^*}=16.65$, the policy $\widehat{\pi}_{\alpha=0.5}$ learns the expert policy with some learning error. As observed in Fig. \ref{fig: learned_policy_nonlinear}, the Lipschitz
constant constraints the slope of the learned surface, where $\widehat{\pi}_{\alpha=0.5}$ has smaller slope than $\widehat{\pi}_{\alpha=50}$, and hence more robust to perturbations in the states. However, since $\widehat{\pi}_{\alpha=0.5}$ has larger learning error, it has poorer nominal performance. Fig. \ref{fig: stabilization} shows the trajectory of the point $(r_t^x, r_t^y)$ (see Fig. \ref{fig: nonholonomic}) induced by the expert policy, $\widehat{\pi}_{\alpha=50}$, and $\widehat{\pi}_{\alpha=0.5}$ starting from initial position $(1,1)$ and an orientation $\theta=180^{\circ}$. The policies are deployed in nominal conditions, Fig. \ref{fig: stabilization}(a), and in non-nominal conditions with $\zeta_r=0.7$ for the position and $\zeta_{\theta}=\pi/180$ for the orientation, Fig. \ref{fig: stabilization}(b). We observe in Fig. \ref{fig: stabilization}(a) that $\widehat{\pi}_{\alpha=50}$ performs as good as the expert and better than $\widehat{\pi}_{\alpha=0.5}$ in nominal conditions. On the other hand, we observe in Fig. \ref{fig: stabilization}(b) that the performance of $\widehat{\pi}_{\alpha=50}$ and that of the expert degrade when deployed in non-nominal conditions, while the performance of $\widehat{\pi}_{\alpha=0.5}$ remains almost the same.
\section{Conclusion}
In this paper propose a framework to learn feedback control policies
with provable robustness guarantees. Our approach draws from our
earlier work~\cite{VK-AAALM-FP:20} where we formulate the
adversarially robust learning problem as one of Lipschitz-constrained
loss minimization.  We adapt this framework to the problem of learning
robust feedback policies from a dataset obtained from expert
demonstrations.  We establish robust stability of the closed-loop
system under the learned feedback policy. 
Further, we derive upper bounds on the regret and
robustness of the learned feedback policy, which bound its nominal
suboptimality with respect to the expert policy and the deterioration
of its performance under bounded (adversarial) disturbances to state measurements,
respectively.  
The above bounds suggest the existence of a tradeoff
between nominal performance of the feedback policy and closed-loop
robustness to adversarial perturbations on the feedback. This tradeoff
is also evident in our numerical experiments, where improving
closed-loop robustness leads to a deterioration of the nominal
performance.  
Finally, we demonstrate our results and the
effectiveness of our robust feedback policy learning framework on 
several benchmarks. 
%
%

%
\bibliographystyle{unsrt}
\bibliography{alias,Main,FP,New}

\section*{Appendix}
\renewcommand{\thesection}{A}

\subsection{Proof of Lemma~\ref{lemma:quad_bound}}
We first note that $V^*(x) \geq 0$ for any $x \in \real^n$, and
$\pi^*(0) = 0$. To see this, we first recall that
$V^*(0) \leq \sum_{t=0}^\infty \gamma^t c(x_t, u_t)$ for $x_{t+1} = f(x_t,u_t)$
with $x_0 = 0$ and any $\{ u_t \}$. In particular, 
with $u_t = 0$ for all $t \in \mathbb{N}$, we get
that $\sum_{t=0}^\infty \gamma^t c(x_t, u_t) = 0$
(since $x_0 = 0$, $f(0,0) = 0$ and $c(0,0) = 0$), 
and since $0 \leq V^*(0) \leq \sum_{t=0}^\infty \gamma^t c(x_t, u_t) = 0$,
it follows that $V^*(0) = 0$.
Now, we have that $\pi^*(0) \in \arg \min_{u \in \real^m} c(0,u) + V^*(f(0,u))$,
from which we clearly get that $\pi^*(0) = 0$ is the only minimizer.
Now, since~$c$ is $\mu$-strongly convex, with $c(0,0) = 0$, we have
$c_{\pi^*}(x) \geq \mu \| x \|^2 / 2$.
Furthermore, since $V^*(x) = \min_{u \in \real^m}	c(x,u) + \gamma V^*(f(x,u))$,
we get:
\begin{align*}
	V^*(x) &= c_{\pi^*}(x) + \gamma V^*(f_{\pi^*}(x))\\ 
				&\geq \frac{\mu}{2} \| x \|^2 + \gamma V^*(f_{\pi^*}(x)) 
				\geq \frac{\mu}{2} \| x \|^2. 
\end{align*}
Let $\pi$ be a Lipschitz-continuous (with constant~$\alpha$), 
exponentially stabilizing feedback policy as in Assumption~\ref{ass:system}.
We then have:
\begin{small}
\begin{align*}
	&V^*(x) \leq V_\pi(x) = \sum_{t=0}^\infty \gamma^t c_{\pi} \left( f_{\pi}^t(x) \right)
				\leq \frac{\lambda}{2} \sqrt{1 + {\alpha}^2} \sum_{t=0}^\infty \gamma^t \left\| f_{\pi}^t(x) \right\|^2 \\
				&\leq \frac{\lambda}{2} \sqrt{1 + {\alpha}^2} \sum_{t=0}^\infty \left\| f_{\pi}^t(x) \right\|^2 
				\leq \frac{\lambda}{2} M \sqrt{1 + {\alpha}^2} \sum_{t=0}^\infty \beta^{2t} \| x \|^2 \\
				&= \frac{\lambda M \sqrt{1 + {\alpha}^2}}{2(1- \beta^2)} \| x \|^2.	
\end{align*}
\end{small}
We then have:
\begin{align*}
	\frac{\mu}{2} \| x \|^2
	\leq V^*(x)
	\leq \frac{\lambda M \sqrt{1 + {\alpha}^2}}{2(1- \beta^2)} \| x \|^2,
\end{align*}
and the statement of the lemma follows.
%

\subsection{Proof of Theorem~\ref{thm:stability}}  	\label{app:robust_stability}
\emph{(i) Exponential stability under expert (optimal) feedback policy:}
We now recall that $V^*(x) = c_{\pi^*}(x) + \gamma V^* \left( f_{\pi^*}(x) \right)$
and $\underline{\kappa}^* \| x \|^2 \leq V^*(x) \leq \bar{\kappa}^* \| x \|^2$ (from Lemma~\ref{lemma:quad_bound}).
It then follows that:
\begin{align*}
	 &V^*(f_{\pi^*}(x)) - V^*(x) \leq - \frac{\mu}{2 \gamma}  \| x \|^2 + \frac{\bar{\kappa}^* (1-\gamma)}{\gamma} \| x \|^2 \\
		&= - \frac{\bar{\kappa}^*}{\gamma} \left[ \gamma - \left( 1 - \frac{\mu}{2 \bar{\kappa}^*} \right) \right] \| x \|^2 
		= - \bar{\kappa}^* \left( 1 - \frac{\gamma'}{\gamma} \right) \| x \|^2.
\end{align*}
It follows from the above inequality and the quadratic boundedness of~$V^*$
that $f_{\pi^*}$ is uniformly globally exponentially convergent~\cite{DT-BR-CK:18}.
In what follows, we obtain an estimate for the upper bound on $\left\| f_{\pi^*}^t(x) \right\|$.
From the above inequality and the fact that $V^*(x) \leq \bar{\kappa}^* \| x \|^2$ we get 
$V^*(f_{\pi^*}(x)) - V^*(x) \leq - (1 - \gamma' / \gamma) V^*(x)$
and $V^*(f_{\pi^*}(x)) \leq \gamma' / \gamma V^*(x)$.
It then follows that $V^*(f_{\pi^*}^t(x)) \leq (\gamma' / \gamma)^t V^*(x)$ which implies
$\underline{\kappa}^* \left\| f_{\pi^*}^t(x) \right\|^2 \leq \bar{\kappa}^* (\gamma'/\gamma )^t \| x \|^2$.
Therefore, we get:
\begin{align*}
	\left\| f_{\pi^*}^t(x) \right\| \leq \sqrt{\frac{\bar{\kappa}^*}{\underline{\kappa}^*}} \left( \sqrt{\frac{\gamma'}{\gamma}} \right)^t \| x \|.
\end{align*}
\medskip
\noindent
\emph{(ii) Robust stability under learned policy:} 
For $x \in B_r(0)$, let $\widehat{x}_t = f_{\widehat{\pi}}^t(x)$ 
and $x_t^* = f_{\pi^*}^t(x)$.
We have:
\begin{align*}
	&\left\| f_{\widehat{\pi}}(x) \right\| \leq \left\| f_{\pi^*}(x) \right\| +  \left\|  f_{\widehat{\pi}}(x) -  f_{\pi^*}(x) \right\| \\
					&\leq \sqrt{\frac{\bar{\kappa}^* \gamma'}{\underline{\kappa}^* \gamma}} \| x \| + \ell_{f}^u  \left\| \widehat{\pi}(x) - \pi^*(x) \right\| 
					\leq \sqrt{\frac{\bar{\kappa}^* \gamma'}{\underline{\kappa}^* \gamma}} r + \ell_{f}^u \varepsilon.
\end{align*}
We see that $\left\| f_{\widehat{\pi}}(x) \right\| \leq r$
for $\varepsilon \leq \frac{1}{\ell_{f}^u} \left( 1 - \sqrt{\frac{\bar{\kappa}^* \gamma'}{\underline{\kappa}^* \gamma}} \right) r$.
It then follows that $\left\| \widehat{\pi}(\widehat{x}_t) - \pi^*(\widehat{x}_t) \right\| \leq \varepsilon$
for any $t \in \mathbb{N}$.
We then have:
\begin{align*}
	\left\| \widehat{x}_t \right\| 
				&\leq  \left( \frac{\bar{\kappa}^* \gamma'}{\underline{\kappa}^* \gamma} \right)^{t/2} \| x \|
									+ \ell_{f}^u \sum_{\tau = 0}^{t-1} \left( \frac{\bar{\kappa}^* \gamma'}{\underline{\kappa}^* \gamma} \right)^{(t - \tau - 1)/2} \varepsilon \\
				&= 	\left( \sqrt{\frac{\bar{\kappa}^* \gamma'}{\underline{\kappa}^* \gamma}} \right)^t \| x \|
									+ \ell_{f}^u  \left[ \frac{1 - \left( \sqrt{\frac{\bar{\kappa}^* \gamma'}{\underline{\kappa}^* \gamma}} \right)^t}{1 - \sqrt{\frac{\bar{\kappa}^* \gamma'}{\underline{\kappa}^* \gamma}}} \right] \varepsilon.			
\end{align*}
Furthermore, we have from part (i) that $x_t^* \in B_r(0)$ for any $t \in \mathbb{N}$.
Therefore, $B_r(0)$ is invariant under $f_{\pi^*}$ and $f_{\widehat{\pi}}$,
and we immediately obtain the uniform bound~$\left\| \widehat{x}_t - x_t^* \right\|  \leq 2r$.
We now have:
\begin{align*}
	&\left\| \widehat{x}_t - x_t^* \right\| 
			= \left\| f(\widehat{x}_{t-1}, \widehat{\pi}(\widehat{x}_{t-1})) - f(x_{t-1}^* , \pi^*(x_{t-1}^*)) \right\| \\
	&\leq \left\| f(\widehat{x}_{t-1}, \widehat{\pi}(\widehat{x}_{t-1})) - f(\widehat{x}_{t-1}, \pi^*(\widehat{x}_{t-1})) \right\|  \\
	&\quad + \left\| f(\widehat{x}_{t-1}, \pi^*(\widehat{x}_{t-1})) - f(x_{t-1}^*, \pi^*(x_{t-1}^*)) \right\| \\
	&\leq \left\| f_{\pi^*}(\widehat{x}_{t-1}) - f_{\pi^*}(x_{t-1}^*) \right\|\\ 
				&\quad+ \left\| f(\widehat{x}_{t-1}, \widehat{\pi}(\widehat{x}_{t-1})) - f(\widehat{x}_{t-1}, \pi^*(\widehat{x}_{t-1})) \right\| \\
	&\leq \ell_{f_{\pi^*}} \left\| \widehat{x}_{t-1} - x_{t-1}^* \right\| + \ell_{f}^u \left\| \widehat{\pi}(\widehat{x}_{t-1}) - \pi^*(\widehat{x}_{t-1}) \right\| \\
	&\leq \ell_{f_{\pi^*}} \left\| \widehat{x}_{t-1} - x_{t-1}^* \right\| + \ell_{f}^u \varepsilon \\
	&\leq \ell_{f}^u \sum_{\tau = 0}^{t-1} \ell_{f_{\pi^*}}^{t-\tau-1} \varepsilon 
	= \ell_f^u \left[ \frac{1 - \ell_{f_{\pi^*}}^t}{1- \ell_{f_{\pi^*}}} \right] \varepsilon,
\end{align*}
where the final equality holds for~$\ell_{f_{\pi^*}} \neq 1$.
If $\ell_{f_{\pi^*}} = 1$, then we have:
\begin{align*}
	\left\| \widehat{x}_t - x_t^* \right\|  \leq t \cdot \ell_f^u \varepsilon.
\end{align*}
We also have:
\begin{align*}
	\left\| \widehat{x}_t - x_t^* \right\| &\leq \left\| \widehat{x}_t \right\| + \left\| x_t^* \right\| \\
					&\leq 2 \left( \sqrt{\frac{\bar{\kappa}^* \gamma'}{\underline{\kappa}^* \gamma}} \right)^t \| x \|
									+ \ell_{f}^u  \left[ \frac{1 - \left( \sqrt{\frac{\bar{\kappa}^* \gamma'}{\underline{\kappa}^* \gamma}} \right)^t}{1 - \sqrt{\frac{\bar{\kappa}^* \gamma'}{\underline{\kappa}^* \gamma}}} \right] \varepsilon.			
\end{align*}
%
%
%
Now, for the policy~$\widehat{\pi}_\delta$, we have $\left\| \widehat{\pi}_\delta - \pi^* \right\|_{\left(B_r(0), \infty \right)}
\leq \left\| \widehat{\pi}_\delta - \widehat{\pi} \right\|_{\left(B_r(0), \infty \right)}
+ \left\| \widehat{\pi} - \pi^* \right\|_{\left(B_r(0), \infty \right)}
\leq \alpha \zeta + \varepsilon$,
and the earlier analysis now carries through with this bound,
and the statement of the theorem follows.

\subsection{Proof of Lemma~\ref{lemma:inc_exp_stability}}
From the exponential stability of~$f_\pi$ 
and $f_\pi$-invariance of $B_r(0)$, for $x, x' \in B_r(0)$,
we have $\left\| f_\pi^t(x) - f_\pi^t(x') \right\| 
\leq \left\| f_\pi^t(x) \right\| + \left\| f_\pi^t(x') \right\|
\leq 2M \beta^t r$.
Furthermore, let~$\ell_{f_\pi}$ be the Lipschitz constant of
$f_\pi$ on~$B_r(0)$. This implies that~$\left\| f_\pi^t(x) - f_\pi^t(x') \right\| 
\leq \ell_{f_\pi}^t \left\| x - x' \right\|$.
We then have $\left\| f_\pi^t(x) - f_\pi^t(x') \right\| 
\leq \min \left \lbrace \ell_{f_\pi}^t \left\| x - x' \right\| ,
2M \beta^t r \right \rbrace$. We therefore obtain an~$M(x_1,x_2)$
such that $\left\| f_\pi^t(x) - f_\pi^t(x') \right\| \leq M(x_1,x_2) \beta^t \| x_1 - x_2 \|$.

\subsection{Proof of Theorem~\ref{thm:regret_robustness_bound}} \label{app:regret_robustness_bound_proof}
The following lemma establishes a difference bound
for the value function under a Lipschitz feedback policy:
\begin{lemma}[\bf \emph{Value function difference bound}]
\label{lemma:lipschitz_trajectory_cost}
  Let $\pi \in \Lip \left( B_r(0); \real^m \right)$ 
  be a Lipschitz feedback policy such that $\pi(0) = 0$, $\lip(\pi) \leq \alpha$. 
  For the value function $V_{\pi}(x) = \sum_{t=0}^\infty \gamma^t c_\pi \left( f_\pi^t(x) \right)$ 
  of policy~$\pi$, the following holds:
  \begin{align*}
  		\left| V_{\pi}(x') - V_{\pi}(x) \right| 
  		\leq 
  		\Theta \lambda r \sqrt{1+\alpha^2} \left\| x' - x \right\|  \left[ 1+ \frac{\left\| x' - x \right\| }{2r} \right],
  \end{align*}
where $\Theta = \sum_{t=0}^{\infty} \gamma^t \theta_t^2$ and $\theta_t = M \beta^t$.
\end{lemma}
\begin{proof}
We first note that $(0,0) \in B_r(0) \times \real^m$ is a strict minimizer of~$c$
(by Assumption~\ref{ass:task}) and since $c$ is differentiable, we have
$\nabla c(0,0) = 0$. 
For any $x \in B_r(0)$:
\begin{align*}
	&\left\| \nabla c_\pi(x) \right\| 
	= \left\| \nabla c(x, \pi(x)) - \nabla c(0,0) \right\| \\
	 &\leq \lambda \left\| (x, \pi(x)) \right\| 
	  \leq \lambda \sqrt{1 + \alpha^2} \| x\|,
\end{align*}
since $\| \pi(x) \| = \| \pi(x) - \pi(0) \| \leq \alpha \| x \|$.
For any $x, x' \in B_r(0)$,
let~$p$ be the straight line segment
between~$x$ and~$x'$, such that 
$p(t) = x + t(x'-x)$ for $t \in [0,1]$.
From the $\lambda$-smoothness of $c$, 
we have:
\begin{align*}
	c_\pi(x') &- c_\pi(x) = \int_0^1 \nabla c_\pi (p(t)) \cdot \dot{p}(t) dt \\
								   &\leq  \int_0^1 \left\| \nabla c_\pi (p(t)) \right\| dt \cdot \| x' - x \| \\
								   &\leq \lambda \sqrt{1 + \alpha^2} \int_0^1 \| p(t) \| dt \cdot \| x' - x \| \\
								   &\leq \lambda \sqrt{1 + \alpha^2} \int_0^1 \| x + t(x'-x) \| dt \cdot \| x' - x \|  \\
								   &\leq \lambda \sqrt{1+\alpha^2} ~\| x \| \left\| x' - x \right\| + \frac{\lambda}{2} \sqrt{1+\alpha^2} \left\| x' - x \right\|^2
\end{align*}
We also have:
\begin{align*}
	c_\pi(x) - c_\pi(x') \leq & \lambda  \sqrt{1+\alpha^2} ~\| x' \| \left\| x' - x \right\| \\
	&+ \frac{\lambda}{2} \sqrt{1+\alpha^2} \left\| x' - x \right\|^2,
\end{align*}
and therefore we get:
\begin{align*}
	\left| c_\pi(x) - c_\pi(x') \right| \leq &\lambda \sqrt{1+\alpha^2}~ \max\{ \| x \|, \| x' \| \} \left\| x' - x \right\| \\
	&+\frac{\lambda}{2} \sqrt{1+\alpha^2} \left\| x' - x \right\|^2.
\end{align*}
 We now have: 
  \begin{align*}
  \footnotesize
  \begin{aligned}
    &V_\pi(x')- V_\pi(x) = \sum_{t=0}^{\infty} \gamma^t \left[ c_{\pi}\left( f_{\pi}^t (x') \right)-c_{\pi}\left( f_{\pi}^t (x) \right) \right] \\
    &\leq \sum_{t=0}^{\infty} \gamma^t \bigg[ \lambda \sqrt{1+\alpha^2}~ \max\{ \| f_{\pi}^t (x) \|, \| f_{\pi}^t (x') \| \} \left\| f_{\pi}^t (x') - f_{\pi}^t (x) \right\| \\
    &\qquad \qquad ~+ \frac{\lambda}{2} \sqrt{1+\alpha^2} \left\|  f_{\pi}^t (x') -  f_{\pi}^t (x) \right\|^2  \bigg] \\
     &\leq \left[ \sum_{t=0}^{\infty} \gamma^t \theta_t^2 \right] \lambda  \sqrt{1+\alpha^2}~ \max\{ \| x \|, \| x' \| \} \left\| x' - x \right\|\\
     	&\quad+ \left[ \sum_{t=0}^{\infty} \gamma^t \theta_t^2 \right] \frac{\lambda}{2} \sqrt{1+\alpha^2} \left\| x' -x \right\|^2 \\
     &\leq \Theta \lambda\sqrt{1+\alpha^2}~ r \left\| x' - x \right\| 
     							+ \frac{1}{2} \Theta \lambda \sqrt{1+\alpha^2} \left\| x' -x \right\|^2,
  \end{aligned}
  \end{align*}
and the statement of the lemma follows.
\end{proof}
\emph{(i) Regret:}
  Let~$\pi \in \Lip \left( B_r(0), \real^m \right)$ be a policy such that 
  $\left\| \pi - \pi^* \right\| \leq \varepsilon$ and
  $\lip(\pi) \leq \alpha$. 
  Since $\varepsilon \leq \frac{1}{\ell_{f}^u} \left( 1 - \sqrt{\frac{\bar{\kappa}^* \bar{\gamma}}{\underline{\kappa}^* \gamma}} \right) r$, 
  we get from~Theorem~\ref{thm:stability} that $B_r(0)$ is $f_\pi$-invariant.
  The value function $V_\pi$ corresponding to~$\pi$
  satisfies $V_\pi(x) = c_\pi \big(x \big) + \gamma V_\pi \big( f_\pi(x) \big)$. 
  We then have for any~$x \in B_r(0)$:
  \begin{align*}
    &\mathcal{R} (\pi) = \sup_{x \in B_r(0)} \left \lbrace V_\pi(x) - V^*(x) \right \rbrace \\
    &= \sup_{x \in B_r(0)} \left \lbrace c_\pi(x) - c_{\pi^*}(x)  + \gamma \left( V_\pi (f_\pi(x)) - V^*(f_{\pi^*}(x)) \right) \right \rbrace \\	
    &\leq \sup_{x \in B_r(0)} \left \lbrace c_\pi(x) - c_{\pi^*}(x) + \gamma \left( V_\pi (f_\pi(x)) -  V^*(f_\pi(x)) \right) \right. \\
						&\qquad \qquad  \left. + \gamma \left( V^*(f_\pi(x)) - V^*(f_{\pi^*}(x)) \right) \right \rbrace \\
	&\leq  \sup_{x \in B_r(0)} \left \lbrace c_\pi(x) - c_{\pi^*}(x)  + \gamma \left( V^*(f_\pi(x)) -  V^*(f_{\pi^*}(x)) \right) \right.\\
		&\qquad \qquad +  \gamma \sup_{x \in B_r(0)} \left \lbrace  V_\pi(x) - V^*(x) \right \rbrace  \rbrace .		
  \end{align*}
 It then follows that:
 \begin{align*}
 	&\mathcal{R} (\pi) \leq \frac{1}{1 - \gamma} \cdot \sup_{x \in B_r(0)} \{ c_\pi(x) - c_{\pi^*}(x) \\
	& \qquad \qquad \qquad \qquad \qquad \; + \gamma \left( V^*(f_\pi(x)) -  V^*(f_{\pi^*}(x)) \right) \}.
 \end{align*}
Furthermore, we have:
\begin{align*}
	&\sup_{x \in B_r(0)} \lbrace c_\pi(x) - c_{\pi^*}(x) \rbrace\\
	 &~\leq \lambda \sqrt{1 + \left| \max \left \{ \alpha , \alpha^* \right \} \right|^2 }  ~r \left\| \pi - \pi^* \right\|_{\infty} + \frac{\lambda}{2} \left\| \pi - \pi^* \right\|_{\infty}^2.
\end{align*}
From Lemma~\ref{lemma:lipschitz_trajectory_cost}, we also have:
\begin{align*}
\small
\begin{aligned}
	&V^*(f_\pi(x)) -  V^*(f_{\pi^*}(x))\\ 
	&\qquad \quad \leq \Theta \lambda r \sqrt{1+{\alpha^*}^2} \left\| f_\pi(x) - f_{\pi^*}(x) \right\|  \left[ 1+ \frac{\left\| f_\pi(x) - f_{\pi^*}(x) \right\| }{2r} \right] \\
	&\qquad \quad \leq \Theta \lambda r \sqrt{1+{\alpha^*}^2}~  \ell_{f}^u \left\| \pi - \pi^* \right\|_{\infty} \left[ 1+ \frac{\ell_{f}^u \left\| \pi - \pi^* \right\|_{\infty}}{2r} \right].
\end{aligned}
\end{align*}
The statement of the theorem follows from the above two inequalities.\\\\
\noindent
\emph{(ii) Robustness:}
    The value function for the policy $\pi$ satisfies
    $V_\pi(x) = c_\pi \big(x \big) + \gamma V_\pi \big(f_\pi(x)
    \big)$.  For convenience of notation, we denote
    $\pi \circ(\id +\delta)$ by $\pi_{\delta}$, i.e.,
    $\pi_{\delta}(x)=\pi(x+\delta)$. For $\|\delta\|_{\infty} \leq \zeta \in \mathbb{R}$, we have:
\begin{align*}
   &\mathcal{S}(\pi) = \sup_{x \in B_r(0)} \left \lbrace V_{\pi_{\delta}} \big(x \big) - V_\pi \big(x \big) \right \rbrace \\
  	&= \sup_{x \in B_r(0)} \left \lbrace c_{\pi_{\delta}}(x) - c_\pi(x) +\gamma \left( V_{\pi_{\delta}} \big(f_{\pi_{\delta}}(x) \big) - V_\pi \big(f_\pi(x) \big) \right) \right \rbrace \\
    &\leq \sup_{x \in B_r(0)} \left \lbrace c_{\pi_{\delta}}(x) - c_\pi(x) + \gamma \left( V_{\pi_{\delta}} \left(f_{\pi_{\delta}}(x) \right) - V_\pi \left(f_{\pi_{\delta}}(x) \right)\right) \right. \\ 	
	&\qquad \qquad \left. +\gamma \left( V_\pi \left(f_{\pi_{\delta}}(x) \right) - V_\pi \left( f_\pi(x) \right)\right) \right \rbrace \\
	&\leq \left. \sup_{x \in B_r(0)} \right \lbrace c_{\pi_{\delta}}(x) - c_\pi(x) + \gamma \left( V_\pi \left(f_{\pi_{\delta}}(x) \right) - V_\pi \left( f_\pi(x) \right) \right) \\ 	
	&\qquad \qquad \left. +\gamma \sup_{x \in B_r(0)} \left \lbrace  V_{\pi_{\delta}} \left(f_{\pi_{\delta}}(x) \right) - V_\pi \left(f_{\pi_{\delta}}(x) \right) \right \rbrace \right \rbrace.
\end{align*}
It then follows that:
\begin{align*}
	&\mathcal{S}(\pi) = \frac{1}{1-\gamma} \cdot \sup_{x \in B_r(0)}  \{ c_{\pi_{\delta}}(x) - c_\pi(x)\\ 
	&\qquad \qquad \qquad \qquad\qquad+ \gamma \left( V_\pi \left(f_{\pi_{\delta}}(x) \right) - V_\pi \left( f_\pi(x) \right) \right)  \}.
\end{align*}
Furthermore, we have:
\begin{align*}
	\sup_{x \in B_r(0)} \lbrace c_{\pi_{\delta}}(x) - c_\pi(x) \rbrace \leq \lambda \alpha \sqrt{1 + \alpha^2 }  ~r \zeta + \frac{\lambda}{2} \alpha^2 \zeta^2.
\end{align*}
From Lemma~\ref{lemma:lipschitz_trajectory_cost}, we also have:
\begin{align*}
\small
\begin{aligned}
	 &V_\pi \left(f_{\pi_{\delta}}(x) \right) - V_\pi \left( f_\pi(x) \right)\\
	&\qquad \quad\leq \Theta \lambda r \sqrt{1+\alpha^2}  \left\| f_{\pi_{\delta}}(x) - f_\pi(x) \right\|  \left[ 1+ \frac{\left\| f_{\pi_{\delta}}(x) - f_\pi(x) \right\|}{2r} \right] \\
	&\qquad \quad\leq \Theta \lambda r \sqrt{1+\alpha^2}~ \ell_{f}^u \alpha \zeta \left[ 1+ \frac{\ell_{f}^u \alpha \zeta}{2r} \right].
\end{aligned}
\end{align*}
The statement of the theorem follows from the above two inequalities.


\subsection{Proof of Theorem~\ref{thm:finite_sample_bound}} 	\label{app:finite_sample_bound_proof}
We first let $X_N = \lbrace \mathbf{x}^{(1)}, \ldots, \mathbf{x}^{(N)} \rbrace$
where $\mathbf{x}^{(i)} = \left( x^{(i)}_0, \ldots, x^{(i)}_{T-1} \right)$ 
and $\rho(X_N, r)$ the covering radius for the set~$X_N$ w.r.t. the ball~$B_r(0)$,
defined as follows:
\begin{align*}
	\rho(X_N, r) = \sup_{x \in B_r(0)} ~\min_{\substack{i \in \lbrace 1, \ldots, N \rbrace, \\ t \in \lbrace 0, \ldots, T-1 \rbrace}} \left| x - x^{(i)}_t \right|.	
\end{align*}

\medskip
\noindent
\emph{(a) From Lipschitz bound to covering radius:}
For any $i \in \lbrace 1, \ldots, N \rbrace$ and $ t \in \lbrace 0, \ldots, T-1 \rbrace$, 
we have:
\begin{small}
\begin{align*} 
	& | \widehat{\pi}(x) - \pi^*(x) | \\
	&=  \left| \widehat{\pi}(x) - \widehat{\pi}(x^{(i)}_t) + \widehat{\pi}(x^{(i)}_t) - \pi^*(x^{(i)}_t) + \pi^*(x^{(i)}_t) - \pi^*(x) \right|  \\
	&\leq (\alpha + \alpha^*) \left| x - x^{(i)}_t \right| + \left| \widehat{\pi}(x^{(i)}_t) - \pi^*(x^{(i)}_t) \right|.
\end{align*}
\end{small}
In particular, the following holds:
\begin{small}
\begin{align*}
	&|\widehat{\pi}(x) - \pi^*(x) |	\\
	&\leq \min_{\substack{i \in \lbrace 1, \ldots, N \rbrace, \\ t \in \lbrace 0, \ldots, T-1 \rbrace}}  \left[ (\alpha + \alpha^*) \left| x - x_i \right| + \left| \widehat{\pi}(x_i) - \pi^*(x_i) \right| \right] \\
	&\leq (\alpha + \alpha^*)  \min_{\substack{i \in \lbrace 1, \ldots, N \rbrace, \\ t \in \lbrace 0, \ldots, T-1 \rbrace}}  \left| x - x_i \right| + \max_{\substack{i \in \lbrace 1, \ldots, N \rbrace, \\ t \in \lbrace 0, \ldots, T-1 \rbrace}}  \left| \widehat{\pi}(x_i) - \pi^*(x_i) \right|  \\
	&\leq (\alpha + \alpha^*)  \min_{\substack{i \in \lbrace 1, \ldots, N \rbrace, \\ t \in \lbrace 0, \ldots, T-1 \rbrace}}  \left| x - x_i \right| + \varepsilon_{\rm train}.
\end{align*}
\end{small}
where
\begin{align*}
\varepsilon_{\rm train} (\widehat{\pi}) = 
	\max_{\substack{i \in \lbrace 1, \ldots, N \rbrace, \\ t \in \lbrace 0, \ldots, T-1 \rbrace}} \left\| \widehat{\pi} ( x_t^{(i)} ) - \pi^* (x_t^{(i)} ) \right\|.
\end{align*}
Therefore, we get:
\begin{align*}
	\sup_{x \in B_r(0)}& \left| \widehat{\pi}(x) - \pi^*(x) \right|\\
	& \leq  (\alpha + \alpha^*) \sup_{x \in B_r(0)} ~\min_{\substack{i \in \lbrace 1, \ldots, N \rbrace, \\ t \in \lbrace 0, \ldots, T-1 \rbrace}} \left| x - x^{(i)}_t \right|+ \varepsilon_{\rm train}.
\end{align*}
From the previous inequality, we obtain the following probabilistic bound:
\begin{small}
\begin{align*}
	\mathbb{P} \Bigg[ \sup_{x \in B_r(0)} \left| \widehat{\pi}(x) - \pi^*(x) \right| > (\alpha + \alpha^*)  \delta  + \varepsilon_{\rm train}  \Bigg]
	\leq \mathbb{P} \left[ \rho(X_N, r) > \delta \right].
\end{align*}
\end{small}
In what follows, we obtain an upper bound for the quantity 
on the right hand side of the inequality above. \\

\medskip
\noindent
\emph{(b) Probability bound for disjoint intersections:} 
Let $Q_1, \ldots, Q_m$ be disjoint subsets of~$B_r(0)$
and let $E_k$ be the event $Q_k \cap X_N = \emptyset$.
We then have:
\begin{align*}
	\mathbb{P} \left[ \cup_{k=1}^m E_k \right] = \sum_{k=1}^m (-1)^{k+1} \sum_{\left( j_1, \ldots, j_k \right)} \mathbb{P} \left[ E_{j_1} \cap \ldots \cap E_{j_k} \right].
\end{align*}
For any~$k$-tuple $\left( j_1, \ldots, j_k \right)$, the 
event $E_{j_1} \cap \ldots \cap E_{j_k}$ occurs if the
points $x_1, \ldots, x_N$ are in the complement of 
$E_{j_1} \cup \ldots \cup E_{j_k}$.
We then have:
\begin{align*}
	\mathbb{P} \left[ E_{j_1} \cap \ldots \cap E_{j_k} \right] = \left[ 1 - \sum_{i=1}^k \frac{\mathrm{Vol} \left( Q_{j_i} \right) }{\mathrm{Vol} \left( B_r(0) \right) } \right]^N.
\end{align*}
For any $k \in \lbrace 1, \ldots, m \rbrace$, let
$c \leq \mathrm{Vol} \left( Q_k \right) \leq C$.
We then have:
\begin{small}
\begin{align*}
	&{m \choose k} \bigg[ 1 - \frac{k \cdot C}{\mathrm{Vol} \left( B_r(0) \right) } \bigg]^N \\ 
			&\leq	\sum_{\left( j_1, \ldots, j_k \right)} \mathbb{P} \left[ E_{j_1} \cap \ldots \cap E_{j_k} \right] 
			\leq  {m \choose k} \left[ 1 - \frac{k \cdot c}{\mathrm{Vol} \left( B_r(0) \right) } \right]^N. 
\end{align*}
\end{small}
It follows from the above that:
\begin{small}
\begin{align} \label{eq:prob_bound_disjoint_intersection}
	\mathbb{P} \left[ \cup_{k=1}^m E_k \right] \leq \sum_{k=1}^m (-1)^{k+1} {m \choose k} \left[ 1 - \frac{k \cdot c(k)}{\mathrm{Vol} \left( B_r(0) \right) } \right]^N,
\end{align}
\end{small}
where $c(k) = c$ when~$k$ is odd and $c(k) = C$ when~$k$ is even. \\

\medskip
\noindent
\emph{(c) Probabilistic bound for covering radius:}
We now use the bound~\eqref{eq:prob_bound_disjoint_intersection} derived in~\emph{(ii)} to obtain
a probabilistic bound for the covering radius of~$X_N$ w.r.t. $B_r(0)$. 
Let $\mathcal{E}_\delta \subset B_r(0)$ be a maximal set of points in $B_r(0)$
such that for any $z_1, z_2 \in \mathcal{E}_\delta$, we have $d(z_1, z_2) \geq \delta$.
Then, for any $x \in B_r(0) \setminus \mathcal{E}_\delta$, there exists $z \in \mathcal{E}_\delta$
such that $d(x,z) < \delta$ (since $\mathcal{E}_\delta$ is maximal, i.e., if $d(x,z) \geq \delta$ we
will have $x \in \mathcal{E}_\delta$ and we will obtain a contradiction).
We note that $\mathcal{E}_\delta$ is a discrete set with $\left| \mathcal{E}_\delta \right|$ 
points (where $\left| \mathcal{E}_\delta \right|$ is the cardinality of $\mathcal{E}_\delta$).
We note that:
\begin{align*}
	\mathrm{Vol} \left( B_r(0) \right) 
			&\geq \sum_{z_k \in \mathcal{E}_{\delta}} \mathrm{Vol} \left( B_r(0) \cap B_{\delta/2}(z_k) \right) \\
				&\geq \left| \mathcal{E}_{\delta} \right| \cdot \mathrm{Vol} \left( B_r(0) \cap B_{\delta/2}(z \in \partial B_r(0)) \right).
\end{align*} 
We therefore get:
\begin{align*}
	\left| \mathcal{E}_{\delta} \right| \leq \frac{\mathrm{Vol} \left( B_r(0) \right)}{\mathrm{Vol} \left( B_r(0) \cap B_{\delta/2}(z \in \partial B_r(0) ) \right)}.
\end{align*}
Furthermore, if $\rho(X_N, r) > \delta$, there exists an $x \in B_r(0)$ such that
$B_\delta(x) \cap X_N = \emptyset$. Now we can choose $z \in \mathcal{E}_{\delta/2}$
such that $d(x,z) < \delta/2$ and it follows that $B_{\delta/2}(z) \cap X_N = \emptyset$.
Let $E_k$ be the event $X_N \cap B_{\delta/2} (z_k) = \emptyset$, where
$\mathcal{E}_{\delta/2} = \lbrace z_1, \ldots, z_{\left| \mathcal{E}_{\delta/2} \right|} \rbrace$.

Now, we note that the dataset~$X_N$ consists of $N$~closed-loop trajectories
of length~$T$, generated by the optimal policy~$\pi^*$.
Now for any $Q \subseteq B_r(0)$ and any $t \in \lbrace 0, \ldots, T-1 \rbrace$,
we have $x_{t} \in Q$ if and only if $x_{0} \in  \left( f_{\pi^*}^{t} \right)^{-1} Q$.
It then follows that a closed-loop trajectory $(x_{0}, \ldots, x_{T-1})$, 
generated by the optimal policy~$\pi^*$, intersects with~$Q$ if and only if
$x_{0} \in B_r(0) \cap \left(\cup_{t=0}^{T-1} \left( f_{\pi^*}^{t}\right)^{-1} Q \right)$.
We now obtain bounds on: 
\begin{align*}
	\mathrm{Vol} \left( B_r(0) \cap \left(\cup_{t=0}^{T-1} \left( f_{\pi^*}^{t}\right)^{-1} Q \right) \right).
\end{align*}
Owing to contractivity of the closed-loop dynamics~$f_{\pi^*}$,
it attains a maximum for $Q = B_{\delta/2}(0)$. 
We have:
\begin{align*}
	\bar{\kappa}^* \left\| f_{\pi^*}(x) \right\|^2 &\geq V^* \left( f_{\pi^*}(x) \right) = -\frac{1}{\gamma} c_{\pi^*}(x) + \frac{1}{\gamma} V^*(x)\\
	&\geq \frac{\underline{\kappa}^* - \lambda \sqrt{1 + {\alpha^*}^2}/2}{\gamma} \| x \|^2.
\end{align*}
It then follows that:
\begin{align*}
	\left\| f_{\pi^*}^t(x) \right\| \geq \left[ \underbrace{\sqrt{\frac{\underline{\kappa}^* - \lambda \sqrt{1 + {\alpha^*}^2}/2}{\gamma \bar{\kappa}^*}}}_{\nu} \right]^t \| x \|,
\end{align*}
Then, the supremum of the volume is given by
$C = \mathrm{Vol} \left( B_r(0) \cap B_{\delta / 2 \nu^T} (0) \right)$.
Also, it attains a minimum for $Q = B_{\delta}(z \in \partial B_r(0))$,
which is lower bounded by $c = \mathrm{Vol} \left( B_r(0) \cap B_{\delta/2}(z \in \partial B_r(0)) \right)$.
We then have:
\begin{align*}
	\mathbb{P} [ \rho(&X_N, r) > \delta ] 
			 = \mathbb{P} \left[ \exists ~x \in B_r(0) \; : \; B_\delta(x) \cap X_N = \emptyset \right] \\
			&\leq \mathbb{P} \left[ \cup_{k=1}^{\left| \mathcal{E}_{\delta/2} \right|} E_k \right] \\
			&\leq \sum_{k=1}^{\left| \mathcal{E}_{\delta/2} \right|} (-1)^{k+1} {\left| \mathcal{E}_{\delta/2} \right| \choose k} \left[ 1 - \frac{k \cdot c(k)}{\mathrm{Vol} \left( B_r(0) \right) } \right]^{N},
\end{align*}
and the statement of the theorem follows.

\subsection{Expert policy for the~system in~subsection~\ref{subsec: nonholonomic example}} \label{sec: nonholonomic example appendix}
In this subsection, we present more details for the numerical example in subsection \ref{subsec: nonholonomic example}. 
The expert's task is to stabilize the point $r_t=[r^x_t,r^y_t]^{\transpose}$ at $[0,0]^{\transpose}$ with minimal cost \eqref{eq: nonlinear control task}. Knowing that $r^x_t=x_t+d\cos(\theta_t)$ and $r^y_t=y_t+d \sin(\theta_t)$ and using \eqref{eq: nonlinear system}, we can describe the dynamics of $r^x_t$ and $r^y_t$ as 
\begin{align}\label{eq: nonlinear system task}
\underbrace{\begin{bmatrix}
r^x_{t+1}\\
r^y_{t+1}
\end{bmatrix}}_{r_{t+1}}
=\underbrace{\begin{bmatrix}
r^x_{t}\\
r^y_{t}
\end{bmatrix}}_{r_t}
+\underbrace{\begin{bmatrix}
T_s\cos(\theta_t) & -d T_s \sin(\theta_t)\\
T_s\sin(\theta_t) & d T_s \cos(\theta_t)
\end{bmatrix}}_{\mc{R}(\theta_t)}
\underbrace{\begin{bmatrix}
v_t\\
\omega_t
\end{bmatrix}}_{u_t},
\end{align}
Where we assumed that $T_s$ is very small and used the approximation $\sin(T_s\omega_t)\approx T_s\omega_t$ and $\cos(T_s\omega_t)\approx 1$. Let $[v_t, \omega_t]^{\transpose}=\mc{R}^{-1}[\mu^x_t, \mu^x_t]^{\transpose}$, then \eqref{eq: nonlinear system task} is written as
\begin{align}\label{eq: nonlinear system compact}
r_{t+1}=r_t + \mu_t, \quad \text{where}\quad \mu_t=[\mu^x_t, \mu^y_t]^{\transpose}.
\end{align}
To stabilize $r_t$ at $[0,0]^{\transpose}$, we design $\mu_t=-K r_t$, where $K$ is a gain matrix that minimizes \eqref{eq: nonlinear control task}, which can be rewritten~as
\begin{align}\label{eq: nonlinear control task rewritten}
\begin{split}
  & \min_{\mu \in \Lip \left( \real^2 ; \real^2 \right)} \qquad \lim_{T \rightarrow \infty} \sum_{t=0}^{T}\gamma^t \left({r_t}^{\transpose}Qr_t + {\mu_t}^{\transpose}R \mu_t\right),\\
  & \qquad \text{s.t.} \qquad \qquad r_{t+1}=r_t +\mu_t,
  \end{split}
\end{align}
We generate $N$ expert trajectories using \eqref{eq: nonlinear system vector form} with $u_t=-\mc{R}^{-1}(\theta_t) K r_t$, where $K$ is the LQR gain matrix that minimizes \eqref{eq: nonlinear control task rewritten} with $T_s=0.01$, $d=0.15$, $\gamma=0.8$, $Q=I_2$, $R=300I_2$, and $\delta=0$. The generated trajectories are contained in the matrices 
\begin{align*}
E =\left[ \begin{matrix} \mathbf{q}^{(1)} & \ldots & \mathbf{q}^{(N)} \end{matrix} \right], 
		\quad  U =\left[ \begin{matrix} \mathbf{u}^{(1)} & \ldots & \mathbf{u}^{(N)} \end{matrix} \right],
\end{align*}
with $\mathbf{q}^{(i)} = ( q_0^{(i)}, \ldots, q_T^{(i)} )$ and
$\mathbf{u}^{(i)} = ( u_0^{(i)}, \ldots, u_{T-1}^{(i)} )$. Each
trajectory is generated with random initial condition, $[x_0^{ ( i )} , y_0^{ ( i )}]^{\T}\in B_2(0)$ and $\theta_0^{ ( i )}\in B_{\pi}(0)$ for $i=1,\dots,N$. 

\end{document}